\documentclass[journal]{IEEEtran}
\usepackage{amsmath,amsfonts}
\usepackage{algorithmic}
\usepackage{algorithm}
\usepackage{array}
\usepackage[caption=false,font=normalsize,labelfont=sf,textfont=sf]{subfig}
\usepackage{textcomp}
\usepackage{stfloats}
\usepackage{url}
\usepackage{verbatim}
\usepackage{graphicx}
\usepackage{cite}

\usepackage{multirow}
\usepackage{makecell}
\usepackage{pifont}

\usepackage[svgnames]{xcolor}
\usepackage[normalem]{ulem} 
\usepackage{pifont}
\usepackage[hidelinks]{hyperref}
\usepackage{CJKutf8}
\usepackage{threeparttable}
\usepackage{etoolbox}

\newif\ifshowrev
\showrevfalse

\ifshowrev
  
  \newcommand{\deleted}[1]{\textcolor{red}{\sout{#1}}}
\else

  \newcommand{\deleted}[1]{}
\fi

\makeatletter
\def\notetopractitionersname{Note to Practitioners}
\newenvironment{Notetopractitioners}{%
    \normalfont
    \if@twocolumn
      \@IEEEabskeysecsize\bfseries\textit{\notetopractitionersname}---\relax
    \else
      \bgroup\par\addvspace{0.5\baselineskip}\centering\vspace{-1.78ex}\@IEEEabskeysecsize\textbf{\notetopractitionersname}\par\addvspace{0.5\baselineskip}\egroup\quotation\@IEEEabskeysecsize
    \fi\@IEEEgobbleleadPARNLSP
}{%
    \relax\ifCLASSOPTIONconference\vspace{0ex}\else\vspace{1.34ex}\fi\par\if@twocolumn\else\endquotation\fi
    \normalfont\normalsize
}
\makeatother

\makeatletter
\def\@IEEEBIOskipN{1.5\baselineskip}
\expandafter\patchcmd\csname\string\IEEEbiography\endcsname
  {plus 1fil}
  {plus 0fil}
  {}{\PackageWarning{DGS-Net}{Could not adjust IEEEbiography spacing}}
\patchcmd{\IEEEbiographynophoto}
  {\vskip 4\baselineskip plus 1fil minus 0\baselineskip}
  {\vskip 1.5\baselineskip}
  {}{\PackageWarning{DGS-Net}{Could not adjust IEEEbiographynophoto spacing}}
\makeatother

\begin{document}
\title{DGS-Net: Learning Dexterous Grasps from Single-View Point Clouds via a Multi-Object Scene Dataset}
\author{Tao~Geng, Dapeng~Yang,~\IEEEmembership{Senior Member,~IEEE}, Ziwei~Liu, Le~Zhang, Le~Qi, Wangyang~Li, Yi~Ren,~\IEEEmembership{Member,~IEEE}, Shan~Luo,~\IEEEmembership{Senior Member,~IEEE}, and Fenglei~Ni%
\thanks{Manuscript received March 16, 2026; revised June 6, 2026 and July 28, 2026; accepted September 24, 2026. This work was supported in part by the National Natural Science Foundation of China under Grant T2388101, in part by the State Key Laboratory of Robotics and Systems, Harbin Institute of Technology, under Grant SKLRS202408B, and in part by the HIT-CMCC 5G Application Innovation Research Institute under Research Project No. 2025A001. (\textit{Corresponding author: Dapeng Yang.})}%
\thanks{Tao Geng, Dapeng Yang, Ziwei Liu, Le Zhang, Le Qi, Wangyang Li, and Fenglei Ni are with the School of Mechatronics Engineering, Harbin Institute of Technology, Harbin 150001, China (e-mail: 25b308016@stu.hit.edu.cn; yangdapeng@hit.edu.cn; 23s008031@stu.hit.edu.cn; 21b908034@stu.hit.edu.cn; ql85245600@126.com; 20b908026@stu.hit.edu.cn; flni@hit.edu.cn).}%
\thanks{Yi Ren is with Huawei Technologies Co., Ltd., Shenzhen 518129, China (e-mail: even.renyi@huawei.com).}%
\thanks{Shan Luo is with King's College London, London WC2R 2LS, U.K. (e-mail: shan.luo@kcl.ac.uk).}}



\maketitle

\begin{abstract}
Dexterous grasping in multi-object scenes remains a fundamental challenge for robotic manipulation, particularly under single-view perception where occlusion and incomplete geometry introduce significant uncertainty. 
To address this challenge, we propose DGS-Net, an end-to-end dexterous grasp prediction network that directly infers complete high-degree-of-freedom grasp configurations from single-view scene point clouds without relying on explicit object reconstruction or segmentation. By explicitly modeling grasp offsets and pre-grasp configurations, DGS-Net effectively compensates for projection errors and geometric ambiguity caused by partial observations, enabling more accurate and physically consistent grasp execution. 
To support effective training of the proposed model, we develop an efficient two-stage dexterous grasp data generation pipeline that first synthesizes dense object-level grasps and then refines them into physically consistent scene-level grasps. Our dataset provides explicit grasp offset modeling, pre-grasp configurations, and point-cloud-level grasp semantic labels, which guide the network to focus on grasp-relevant regions, thereby enhancing the precision and robustness of grasp prediction under occluded conditions. 
Extensive real-world experiments demonstrate that DGS-Net achieves a grasp success rate of 79.39\% and a clearance rate of 87.94\% on a physical robotic platform. Moreover, compared to the baselines, our method reduces the translation error by approximately 29\% and the rotation error by about 41\% in grasp perturbation tests, validating its effectiveness, robustness, and generalization performance for single-view dexterous grasping in multi-object scenes. 
Our dataset is available at \href{https://github.com/4taotao8/DGS-Net}{https://github.com/4taotao8/DGS-Net}.
\end{abstract}

\begin{Notetopractitioners}
Dexterous grasping is a fundamental capability for robots, and this work aims to address the practical challenges of deploying multi-fingered dexterous hands in unstructured industrial environments such as intelligent manufacturing and automated sorting. Conventional grasping systems often perform poorly in multi-object scenes because a single camera cannot observe object regions that are occluded. To address this limitation, we develop DGS-Net, an end-to-end framework that directly predicts dexterous grasp configurations from single-view scene point clouds without requiring explicit object segmentation or shape reconstruction. To improve grasp reliability under partial observations, the proposed method explicitly models grasp offsets and pre-grasp configurations. These mechanisms help compensate for perception errors caused by occlusions and sensor noise, enabling more accurate dexterous grasping and safer approach motions. In addition, we introduce a cost-effective two-stage data generation pipeline that produces dexterous grasp annotations with semantic labels and grasp offsets, achieving comparable grasping performance with a much smaller dataset, thereby significantly reducing the cost of dataset construction for practical deployment. The current limitation is that the method mainly focuses on the hand itself and is, to some extent, decoupled from the kinematic constraints of the robotic arm. Future work will integrate arm motion planning with grasp prediction, incorporate multimodal sensing such as tactile feedback to adapt grasps to objects' physical properties, and further enrich the diversity of dexterous grasp patterns.
\end{Notetopractitioners}


\section{Introduction}

\IEEEPARstart{G}{rasping} refers to the process of exerting forces and torques at a set of contact points to control an object's motion in a desired manner \cite{newburyDeepLearningApproaches2023}. As a fundamental robotic capability, grasping underpins a wide range of higher-level tasks, including object manipulation, assembly, and physical interaction. In recent years, with the rapid development of humanoid robotics, grasping has attracted increasing attention from both academia and industry and has achieved remarkable progress as a core technology enabling intelligent and versatile robot operation \cite{songOverviewLearningbasedDexterous2025}. Nevertheless, identifying an optimal grasp from an extremely large candidate space remains a challenging problem \cite{bohgDataDrivenGraspSynthesis2014,billardTrendsChallengesRobot2019}. This is mainly due to the high dimensionality of dexterous hand configurations, the numerous possible contact combinations between multi-fingered hands and objects, and the complex, highly coupled interactions among the hand, the object, and the environment.

Early studies on robotic grasping have primarily relied on analytical approaches, which formulate grasp synthesis as an optimization problem grounded in known object geometry and physical properties \cite{bohgDataDrivenGraspSynthesis2014,desouzaRoboticGraspingWrench2021}. Under the assumption that object models can be obtained or that observed irregular objects can be approximated through simplification or abstraction methods \cite{millerAutomaticGraspPlanning2003,goldfederGraspPlanningDecomposition2007,huebnerSelectionRobotPregrasps2008,bergströmIntegrationVisualCues2009}, these methods construct grasp configurations under simplified models of contact mechanics, Coulomb friction, and rigid-body dynamics, and then evaluate grasp quality using criteria such as force closure~\cite{nguyenConstructingForceClosure1988} or form closure~\cite{muellerModernRoboticsMechanics2019}. While analytical methods offer clear physical interpretability and theoretical guarantees, they often depend on idealized assumptions that fail to capture contact uncertainty, perception noise, and environmental interactions in real-world scenarios. Moreover, for multi-fingered dexterous hands, the high dimensionality of the grasp space leads to an exponentially growing number of candidate grasps, resulting in substantial computational overhead \cite{bohgDataDrivenGraspSynthesis2014}. Consequently, these factors greatly limit the generalization capability and efficiency of analytical grasping methods in novel multi-object environments.

With the rapid advancement of artificial intelligence, particularly deep learning and reinforcement learning, data-driven grasping approaches have gained increasing attention due to their effectiveness in robotic grasp prediction. By learning mappings from sensory observations to grasp configurations using large-scale grasping experience \cite{newburyDeepLearningApproaches2023}, these methods enable more flexible and robust grasp predictions for novel objects and unseen environments. 

In data-driven grasp prediction, the choice of end-effector plays a critical role in grasp representation and learning complexity, with parallel grippers and multi-fingered dexterous hands exhibiting fundamentally different characteristics. For parallel grippers, grasp configurations are typically parameterized by a unified 7D representation consisting of 3D translation, 3D rotation, and 1D gripper width. This representation has facilitated large-scale dataset collection \cite{fangGraspNet1BillionLargeScaleBenchmark2020,eppnerACRONYMLargeScaleGrasp2021,backGraspClutter6DLargescaleRealworld2025}, improved generalization across objects and scenes, and fostered a wide range of data-driven grasp prediction methods \cite{fangGraspNet1BillionLargeScaleBenchmark2020,fangAnyGraspRobustEfficient2023,breyerVolumetricGraspingNetwork2021,daiGraspNeRFMultiviewbased6DoF2023}. However, their limited degrees of freedom (DoF) and restricted contact patterns inherently constrain their adaptability to complex grasp tasks. In contrast, multi-fingered dexterous hands (e.g., DLR/HIT Hand II \cite{liuMultisensoryFivefingerDexterous2008}, Shadow Hand \cite{ShadowDexterousHand2023}), as well as soft robotic hands~\cite{zhouCableDrivenSoftRobotic2026}, exhibit human-like morphology with higher DoF, multiple contact points, and rich kinematic redundancy, enabling more flexible and precise grasping behaviors in cluttered environments and for irregular objects \cite{billardTrendsChallengesRobot2019,thalhammerChallengesMonocular6D2024,huangHumanlikeDexterousManipulation2025}. Nevertheless, the diverse mechanical structures and high-dimensional grasp spaces of dexterous hands make it difficult to adopt a unified grasp representation, substantially increasing the complexity of learning and inference. Recently, several studies \cite{casasMultiGripperGraspDatasetRobotic2024,freibergDiffusionMultiEmbodimentGrasping2025} have explored unified grasp learning frameworks across multiple robotic hands to improve generalization across different embodiments. However, existing data-driven methods still struggle to fully exploit the potential of dexterous hands, and current robotic grasping systems remain far from achieving human-level dexterity \cite{huangHumanlikeDexterousManipulation2025}.

Early studies \cite{liuGeneratingGraspPoses2019,gabellieriGraspItPro2020,jiangHandObjectContactConsistency2021} often restrict the problem to isolated objects with complete point clouds or meshes, where grasp prediction is conducted under the assumption of full object geometry. However, obtaining complete 3D representations typically requires multi-view scanning and reconstruction, which is time-consuming and difficult to integrate into real-world robotic systems that demand fast and reactive perception. To alleviate this limitation, recent works \cite{kiatosGeometricApproachGrasping2021,lundellMultiFinGANGenerativeCoarseToFine2021,weiDVGGDeepVariational2022,humtCombiningShapeCompletion2023} attempt to recover complete object geometry from single-view observations by first segmenting target objects using methods such as Segment Anything (SAM) \cite{kirillovSegmentAnything2023} and then applying learning-based shape completion networks. Nevertheless, as pointed out in recent analyses on the readiness of single-view reconstruction for robotics \cite{nolteSingleviewMeshReconstruction2025}, current completion methods \cite{boulchPOCOPointConvolution2022,bossSF3DStableFast2025,xuInstantMeshEfficient3D2024} still suffer from substantial inaccuracies in unseen regions, leading to geometric distortions that can severely degrade downstream grasping and manipulation performance. In particular, grasp prediction based on completed shapes is prone to increased penetration and physical inconsistency due to accumulated reconstruction errors. Moreover, predicting grasps for individual objects in isolation neglects critical scene-level context, often resulting in collision-prone or infeasible grasps. Motivated by these limitations, more recent works \cite{liHGCNetDeepAnthropomorphic2022,zhangDexGraspNet20Learning2024} have shifted toward directly predicting grasps from single-view point clouds of multi-object scenes, without relying on explicit scene reconstruction or single-object shape completion. Building upon this line of research, our goal is to further improve the accuracy and physical reliability of dexterous grasp prediction by directly inferring grasps from single-view scene-level point clouds in an end-to-end manner.

Another key bottleneck lies in the availability of dexterous grasp datasets. Collecting large-scale, high-quality grasp data in real-world settings is notoriously costly and labor-intensive. To mitigate this issue, several recent works have explored the construction of large-scale synthetic dexterous grasp datasets in simulation \cite{zhangDexGraspNet20Learning2024,wangDexGraspNetLargeScaleRobotic2023,chenBODexScalableEfficient2025,chenDexonomySynthesizingAll2025}, which have substantially promoted data-driven learning for dexterous grasp prediction. UltraDexGrasp~\cite{yangUltraDexGraspLearningUniversal2026} further constructs a large-scale synthetic dataset for bimanual dexterous grasping using optimization-based synthesis. However, due to the heterogeneous mechanical designs and varying degrees of freedom across different multi-fingered dexterous hands, these datasets are typically collected for a specific hand model or a small set of dexterous hands, making it difficult to achieve the level of generality observed in parallel-gripper grasp datasets. In practical applications, regenerating large-scale grasp datasets for each new dexterous hand is computationally expensive and time-consuming, posing a substantial barrier to real-world deployment. Therefore, from an application-oriented perspective, it is more desirable to develop efficient dexterous grasp data generation pipelines, together with learning frameworks that can effectively leverage such data.

In this work, we focus on predicting dexterous grasp configurations directly from single-view point clouds in multi-object scenes. Under the challenging conditions of partial observations and sensor noise inherent to single-view perception, our method aims to improve the accuracy and physical reliability of dexterous grasp prediction. The main contributions of this paper are summarized as follows:
\begin{itemize}
\item{We propose DGS-Net, an end-to-end dexterous grasp prediction network that directly predicts dexterous grasp configurations from single-view point clouds in multi-object scenes, enabling more accurate and reliable grasping under partial and noisy observations.}
\item{We propose a cost-effective two-stage dexterous grasp data generation pipeline that generates dense object-level grasps and refines them into collision-free scene-level grasps, enabling efficient generation of point-cloud-level annotations with grasp semantic labels for dexterous grasp prediction.}
\item{We propose learning-based grasp offset prediction to compensate for occlusion and sensor noise in single-view point clouds, enabling more accurate grasp execution under partial observations. In addition, a simple yet effective pre-grasp configuration is introduced to improve robustness in real-world dexterous grasping.}
\item{Real-world experiments demonstrate that DGS-Net achieves a grasp success rate of 79.39\% and a clearance rate of 87.94\% on a physical robotic platform. Moreover, compared to the baselines, our method reduces the translation error by approximately 29\% and the rotation error by about 41\% in grasp perturbation tests.}
\end{itemize}

\section{Related Work}
\subsection{Analytical Grasping}
Analytical grasping methods typically formulate grasp synthesis as an optimization problem under the assumption that object geometry and the kinematic and physical properties of dexterous hands are known \cite{bohgDataDrivenGraspSynthesis2014}. By adopting simplified contact models, Coulomb friction, and rigid-body dynamics, these approaches construct grasp configurations that satisfy force-closure or form-closure conditions to resist external forces and torques \cite{desouzaRoboticGraspingWrench2021}. Among them, GraspIt! \cite{millerGraspitVersatileSimulator2004} is a classical tool that synthesizes stable grasps through collision detection and grasp quality evaluation. However, the high dimensionality of dexterous hand configurations and the combinatorial nature of contact selection lead to an extremely large grasp space, making exhaustive search and evaluation computationally expensive.

To reduce the search complexity, many analytical methods simplify or abstract object geometry. Representative approaches approximate objects using basic geometric primitives \cite{millerAutomaticGraspPlanning2003}, decompose objects into multiple parts represented by bounding boxes \cite{huebnerSelectionRobotPregrasps2008}, or fit local geometry with superquadrics (SQs) \cite{goldfederGraspPlanningDecomposition2007}. While these strategies improve computational tractability, they inevitably sacrifice geometric fidelity.

Another line of work constrains the search space by incorporating human grasping experience. The work \cite{herzogTemplatebasedLearningGrasp2012} uses human-demonstrated grasps as templates and retrieves suitable grasps for novel objects by matching object shape similarity. The work \cite{romeroModelingEvaluationHumantorobot2009} observes human hand postures during grasping and maps them to robotic dexterous hands to generate feasible grasp configurations.

\subsection{Data-Driven Grasping}
With the rapid development of artificial intelligence, particularly deep learning, data-driven grasping methods have attracted increasing attention. Unlike analytical approaches, data-driven methods learn grasping strategies directly from large-scale data. By leveraging deep neural networks, they can implicitly model complex hand-object interactions and high-dimensional grasp configurations, making them well suited for dexterous hands and objects with complex geometry. Consequently, data-driven methods achieve improved generalization and robustness \cite{bohgDataDrivenGraspSynthesis2014,newburyDeepLearningApproaches2023,songOverviewLearningbasedDexterous2025}. Table~\ref{tab:data_driven_grasp} summarizes representative recent data-driven grasp-prediction methods.

\begin{table*}[b]
\caption{A Summary of Data-Driven Grasping Methods}
\label{tab:data_driven_grasp}
\centering
\small
\setlength{\tabcolsep}{2pt}
\begin{tabular}{c|cc|cc|c|c}
\Xhline{1pt}
\multirow{2}{*}{Method} & \multicolumn{2}{c|}{View Setting} & \multicolumn{2}{c|}{Scene Setting} & \multirow{2}{*}{Shape Completion} & \multirow{2}{*}{Input} \\
\cline{2-5}
 & Single-view & Multi-view & Single-object & Multi-object &  &   \\
\Xhline{0.8pt}
\cite{duanLearningHumantoRobotDexterous2023,xieDexMGNetMultiModeDexterous2025,qiuyuchenDexTransferRealWorld2022,blättnerDMFCGraspNetDifferentiableMultiFingered2023,mayerFFHNetGeneratingMultiFingered2022} & \ding{51} & & \ding{51} & & None & Point Cloud \\
\cite{weiDVGGDeepVariational2022,gabellieriGraspItPro2020,palleschiGraspItPro2023,kimDSQNetDeformableModelBased2023,humtCombiningShapeCompletion2023,lundellMultiFinGANGenerativeCoarseToFine2021}& \ding{51} & & \ding{51} & & Object-level & Point Cloud \\
\cite{kiatosGeometricApproachGrasping2021} & \ding{51} & & \ding{51} & & Object-level & SDF \\
\cite{aktaşDeepDexterousGrasping2022,fengDexGANGraspDexterousGenerative2024} & \ding{51} & & \ding{51} & & None & RGB-D \\
\cite{lundellDDGCGenerativeDeep2021}   & \ding{51} & & & \ding{51} & Scene-level & RGB-D \\
\cite{lundellTopgraspsSceneCompletion2020} & \ding{51} & & & \ding{51} & Scene-level & Point Cloud \\
\cite{jiangSynergiesAffordanceGeometry2021}  & \ding{51} & & & \ding{51} & Scene-level & TSDF \\
\cite{fangGraspNet1BillionLargeScaleBenchmark2020,zhangDexGraspNet20Learning2024,weiCMGNetEndtoEndContactbased2023,liHGCNetDeepAnthropomorphic2022,fangAnyDexGraspGeneralDexterous2025,zhangCADGraspLearningContact2025} & \ding{51} & & & \ding{51} & None & Point Cloud \\
\cite{breyerVolumetricGraspingNetwork2021} & \ding{51} & & & \ding{51} & None  & TSDF \\
\cite{burkhardtMultifingeredDynamicGrasping2024,xuUniDexGraspUniversalRobotic2023,wanUniDexGraspImprovingDexterous2023,zhongDexGraspAnythingUniversal2025} & & \ding{51} & \ding{51} & & None & Point Cloud \\
\cite{liuDeepDifferentiableGrasp2020} & & \ding{51} & \ding{51} & & None & Depth Images \\
\cite{liPlanningMultifingeredGrasps2023} & & \ding{51} & & \ding{51} & Scene-level & TSDF \\
Ours   & \ding{51} &   &   & \ding{51}  & None & Point Cloud \\
\Xhline{1pt}
\end{tabular}
\end{table*}

\textbf{Single-object Grasp Prediction:} Some data-driven methods focus on predicting dexterous grasps from geometric representations of isolated objects. For example, \cite{jiangHandObjectContactConsistency2021} generates high-quality human-like grasps by performing test-time self-supervised optimization, while \cite{weiDROGraspUnified2024} takes the robot hand description together with an object point cloud as input and efficiently predicts kinematically valid and stable grasp configurations. More recently, UniGraspTransformer~\cite{wangUniGraspTransformerSimplifiedPolicy2025} distills object-specific reinforcement learning policies into a universal Transformer for scalable and generalized single-object grasping. Benefiting from the availability of complete object geometry, these approaches can explicitly model hand-object interactions, often achieving high grasp precision and stable contact patterns.

Despite their effectiveness, single-object grasp prediction methods face notable challenges in real-world scenarios. First, they typically require isolating individual objects from a scene, followed by grasp prediction in a sequential manner. This object-centric pipeline ignores scene-level context, and the predicted grasps often need to be mapped back into the original scene for additional collision checking, resulting in a time-consuming and inefficient grasp planning pipeline. Second, these methods rely on complete object geometry, which is difficult to obtain for novel objects in real environments and often necessitates an additional shape completion process, thereby increasing system complexity and introducing potential error accumulation that can degrade grasp reliability. Although some works \cite{mayerFFHNetGeneratingMultiFingered2022,blättnerDMFCGraspNetDifferentiableMultiFingered2023,duanLearningHumantoRobotDexterous2023,xieDexMGNetMultiModeDexterous2025} attempt to directly predict grasps from single-view partial point clouds of isolated objects, the resulting grasp quality is typically lower than that achieved by methods operating on complete object geometry. To address this limitation, a number of recent works incorporate shape completion into the grasp prediction process, aiming to compensate for missing geometry during perception.

\textbf{Shape Completion for Grasping:} Shape completion and grasp prediction are closely related tasks, as both require fine-grained understanding of local object geometry \cite{jiangSynergiesAffordanceGeometry2021}. To address incomplete shape perception caused by occlusion, many works incorporate shape completion into the grasp planning pipeline. For example, \cite{liPlanningMultifingeredGrasps2023} reconstructs the Truncated Signed Distance Function (TSDF) of a cluttered scene by fusing depth images from a fixed dual camera system, and then plans multi-fingered grasps on the reconstructed geometry. However, practical robotic systems generally favor grasp prediction directly from single-view observations.

Under the single-view setting, several works attempt to alleviate the incompleteness of partial observations through explicit or implicit shape completion. \cite{gabellieriGraspItPro2020} and its extended version \cite{palleschiGraspItPro2023} perform coarse shape approximation by representing the underlying object geometry with multiple minimum-volume bounding boxes fitted to the observed partial point cloud. \cite{kiatosGeometricApproachGrasping2021} recovers full object geometry by inferring a complete signed distance field (SDF) from incomplete SDF observations. \cite{kimDSQNetDeformableModelBased2023}, \cite{weiDVGGDeepVariational2022}, and \cite{humtCombiningShapeCompletion2023} explicitly reconstruct complete object representations, such as parametric superquadric models, dense point clouds, or meshes, from single-object partial point clouds and subsequently perform grasp prediction based on the reconstructed geometry. \cite{lundellDDGCGenerativeDeep2021} performs scene-level shape completion by reconstructing a completed voxel grid of the scene from a single depth observation, on which dexterous grasps are generated.
\cite{jiangSynergiesAffordanceGeometry2021} and \cite{lundellMultiFinGANGenerativeCoarseToFine2021} further integrate shape completion directly into the grasp generation pipeline, enabling grasp synthesis to be conditioned on implicitly completed object geometry.

Although these approaches alleviate the impact of occlusion to some extent, the accuracy of the completed geometry remains limited, especially for complex shapes, which can propagate errors to downstream grasp prediction. Moreover, explicit shape completion usually introduces additional computational overhead, reducing the efficiency of grasp prediction in practical robotic systems. Recent analyses \cite{nolteSingleviewMeshReconstruction2025} further suggest that current single-view reconstruction methods are still insufficiently reliable for precise robotic manipulation, motivating alternative approaches that bypass explicit shape completion and directly reason about grasping under partial observations.

\textbf{End-to-end Grasp Prediction:} Recent data-driven methods have begun to predict dexterous grasps in an end-to-end manner directly from single-view scene-level observations, avoiding the complexity and potential error accumulation introduced by explicit shape completion. For example, \cite{liHGCNetDeepAnthropomorphic2022} proposes an end-to-end framework that operates on single-view point clouds of multi-object scenes and predicts dexterous grasps by learning grasp-related geometric features in local regions. \cite{zhangDexGraspNet20Learning2024} introduces DexGraspNet 2.0, a large-scale synthetic dataset together with a generative grasping framework, which achieves state-of-the-art performance in scene-level dexterous grasp prediction under cluttered settings. More recently, CADGrasp~\cite{zhangCADGraspLearningContact2025} predicts a contact- and collision-aware intermediate representation to optimize stable dexterous grasps in cluttered scenes.

Despite their effectiveness, end-to-end grasp prediction from single-view scene point clouds remains challenging. Since the input contains many background points and points from non-graspable regions, these methods heavily rely on careful network design and large-scale datasets to guide the model toward grasp-relevant local structures. Moreover, occlusions and sensor noise inherent in single-view perception can significantly degrade grasp quality, often leading to penetration or inaccurate contact placement. In this work, we address these challenges by introducing explicit grasp semantics to guide the network's attention toward regions with high grasp potential, and by proposing a learnable grasp offset to compensate for perception-induced errors. To the best of our knowledge, no prior work has explicitly modeled grasp offset prediction for dexterous grasping, enabling more accurate and reliable grasp execution under partial observations.

\subsection{Grasp Datasets}
High-quality grasp datasets are a prerequisite for data-driven dexterous grasping. \cite{hampaliHOnnotateMethod3D2020} and \cite{chaoDexYCBBenchmarkCapturing2021} provide multi-view RGB-D human grasp datasets, but require additional mapping to be applicable to multi-fingered robotic hands. \cite{wangDexGraspNetLargeScaleRobotic2023,chenBODexScalableEfficient2025} and \cite{chenDexonomySynthesizingAll2025} construct large-scale single-object dexterous grasp datasets in simulation, while RealDex~\cite{liuRealDexHumanlikeGrasping2024} further incorporates real human grasp demonstrations to improve realism and contact fidelity. However, these single-object datasets generally lack scene-level context. \cite{zhangDexGraspNet20Learning2024} takes a key step forward by constructing a large-scale multi-object scene-level dexterous grasp dataset. More recently, HUG~\cite{wuHumanUniversalGrasping2026} introduces a large-scale egocentric human grasp dataset that supports grasp transfer to multiple robotic hands.

Despite recent progress, existing dexterous grasp datasets are typically constructed for a specific hand or a small set of hands. Regenerating large-scale grasp datasets for new dexterous hands or scene configurations is computationally expensive and time-consuming, posing a major obstacle to practical deployment. To address these challenges, we propose a cost-effective two-stage dexterous grasp data generation pipeline, which first generates dense object-level grasps and then efficiently refines them into collision-free scene-level grasps. The proposed pipeline produces point-cloud-level dexterous grasp annotations together with grasp offsets, which are typically not provided by existing datasets, enabling data-efficient dexterous grasp prediction in multi-object scenes.

\section{Problem Statement}
\begin{figure}[t]
\centering
\includegraphics[width=2.8in]{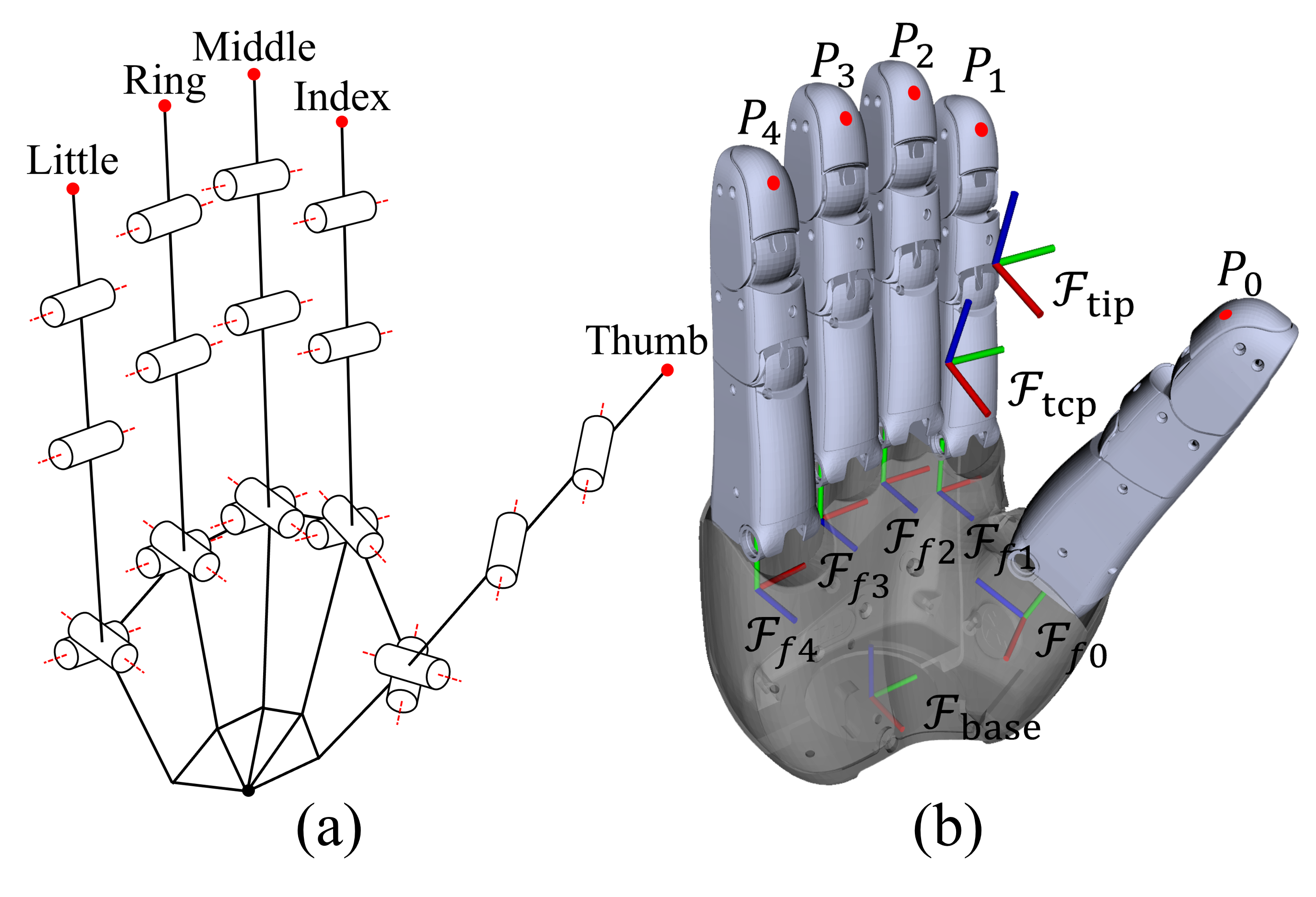}
\caption{(a) Joint configuration. (b) Fingertip points and coordinate systems.}
\label{fig:hand}
\end{figure}

Given a single-view incomplete point cloud of a multi-object scene, we aim to predict dexterous grasps for multiple objects. Our method employs a two-stage grasp prediction network that takes the point cloud as input and outputs grasp poses together with the corresponding joint configurations, including pre-grasp configurations. In this work, we employ the DexHand Pro, a variant of the DLR/HIT Hand II \cite{liuMultisensoryFivefingerDexterous2008}, which features a modular design with five identical fingers, comprising 20 joints and 15 active degrees of freedom, with each finger consisting of four joints, as shown in Fig.~\ref{fig:hand}(a).

Several key definitions are presented here:

\textbf{Hand Grasps:} The grasp configuration of the hand is represented as $G = (T, R, J_{\text{init}}, J_{\text{end}})$. Specifically, $T \in \mathbb{R}^3$ denotes the tool center point (TCP) of the dexterous hand, and $R \in \mathbb{R}^6$ represents the 6D rotation \cite{zhouContinuityRotationRepresentations2019} of the dexterous hand in the camera coordinate frame. $J_{\text{init}}, J_{\text{end}} \in \mathbb{R}^{20}$ correspond to the 20-DoF joint angles of the hand in the pre-grasp and final grasp configurations, respectively.

\textbf{Hand Coordinate Frames:} We define multiple coordinate frames for the hand, including the base frame $\mathcal{F}_{\mathrm{base}}$, the tool center point frame $\mathcal{F}_{\mathrm{tcp}}$, the base frame of each finger $\mathcal{F}_{f_i}$, and the fingertip plane frame $\mathcal{F}_{\mathrm{tip}}$. The frame $\mathcal{F}_{\mathrm{tip}}$ is defined by translating $\mathcal{F}_{\mathrm{tcp}}$ along its $z$-axis such that the sum of vertical distances from all fingertip points $P_i = (x_i, y_i, z_i)$ to the $xy$-plane of $\mathcal{F}_{\mathrm{tip}}$ is minimized. Here, $i=0,1,2,3,4$ denotes the thumb, index, middle, ring, and little fingers, respectively, as illustrated in Fig.~\ref{fig:hand}(b).

\textbf{Point Clouds:} We denote the point cloud of the scene captured by the camera as $\mathcal{P} \in \mathbb{R}^{N \times 3}$, where $N$ is the number of points.

\textbf{Task Description:} Given a scene containing a random number $j$ of objects, a single-view point cloud $\mathcal{P} \in \mathbb{R}^{N \times 3}$ is captured by a depth camera. The proposed end-to-end grasp prediction network, DGS-Net, takes $\mathcal{P}$ as input and predicts the dexterous hand's TCP $T$, hand rotations $R$, and joint configurations $J_{\mathrm{init}}$ and $J_{\mathrm{end}}$ for each object. The robot then executes grasping actions sequentially based on the predicted grasp scores, with the objective of sequentially removing the objects from the scene.

To support this task, we propose a two-stage dataset generation pipeline for constructing training data. In the first stage, a dense set of candidate grasps $\mathcal{G} = \{G_1, G_2, \cdots, G_i\}$ is generated for each individual object using a fingertip constraint strategy, where $i$ denotes the number of feasible grasp configurations. In the second stage, $j$ objects are randomly placed into a shared scene, and the grasp set $\mathcal{G}$ from the first stage is transformed to the scene to form $\tilde{\mathcal{G}} = \{\mathcal{G}_1, \mathcal{G}_2, \cdots, \mathcal{G}_j\}$. We then filter $\tilde{\mathcal{G}}$ based on whether the dexterous hand collides with objects in the scene, resulting in the final scene-level dense grasp set $\mathcal{G}_{\mathrm{scene}}$. In simulation, we render RGB and depth images from multiple viewpoints and generate a set of single-view point clouds $\{\mathcal{P}_k \mid k = 1, 2, 3, \dots\}$, which are paired with $\mathcal{G}_{\mathrm{scene}}$ to construct the training samples $(\mathcal{P}_k, \mathcal{G}_{\mathrm{scene}})$.

\section{Grasp Prediction}
\begin{figure*}[!t]
\centering
\includegraphics[width=7.14in]{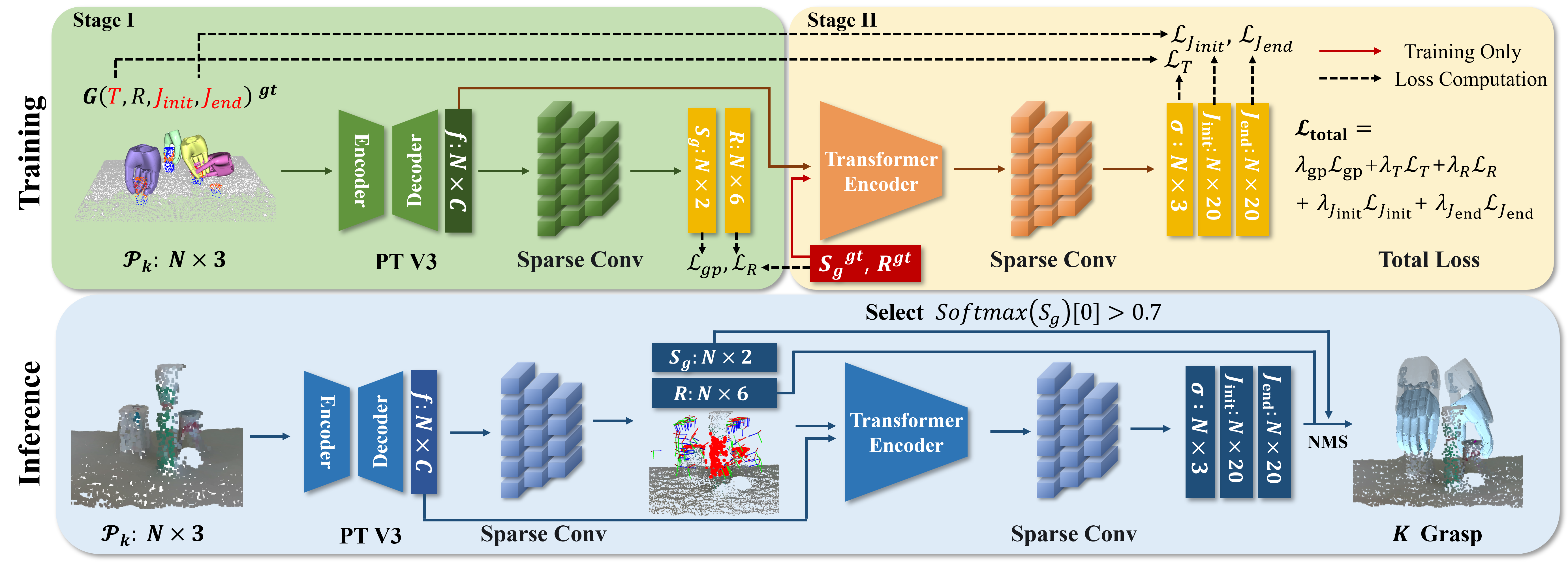}
\caption{Overview of DGS-Net. PTv3 extracts multi-scale point cloud features. Module I (green) predicts grasp reference point and 6D rotations; Module II (yellow) generates grasp offsets and joint configurations. During training (top), ground-truth grasp reference point and pose labels guide Module I to speed convergence and improve stability. During inference (bottom), Module II uses Module I predictions, forming an end-to-end pipeline.}
\label{fig:DGS-Net}
\end{figure*}

In this section, we introduce DGS-Net, an end-to-end network for dexterous grasp prediction. The overall pipeline is illustrated in Fig.~\ref{fig:DGS-Net}. 

Conventional point cloud networks such as PointNet++~\cite{qiPointNetDeepHierarchical2017} mainly rely on hierarchical local neighborhood construction based on $k$-nearest-neighbor (KNN) or ball query operations, which are inherently limited in modeling long-range spatial dependencies and correlations between distant local structures. In contrast, our framework adopts Point Transformer V3 (PTv3)~\cite{wuPointTransformerV32024} as the backbone to efficiently capture global contextual information through spatial serialization and global self-attention mechanisms. Such a design is particularly suitable for dexterous manipulation tasks, where grasp feasibility depends not only on local surface geometry but also on scene-level understanding, including object-object interactions, occlusion relationships, and surrounding obstacles. Benefiting from its large receptive field and multi-scale feature representation capability, PTv3 enables our network to jointly model fine-grained local structures and global scene context for robust dexterous grasp prediction in multi-object environments.

DGS-Net follows a two-stage prediction strategy, as illustrated in Fig.~\ref{fig:DGS-Net}. The first stage, shown in green, serves as a global scene understanding module. It takes the single-view scene point cloud as input and uses PTv3 to extract multi-scale contextual features, based on which it predicts a grasp reference point score and a corresponding 6D grasp rotation representation for each point. The second stage, shown in yellow, serves as a grasp refinement module. It takes the point-wise features from the first stage, together with the grasp reference point scores and grasp rotations, and aggregates local and global geometric information through a self-attention module. Based on these refined features, it predicts the grasp offset, which compensates for the discrepancy between the selected reference point and the dexterous hand TCP, as well as the pre-grasp and final grasp joint configurations.

To facilitate optimization and improve learning stability, additional intermediate supervision is introduced during training. Specifically, the second module receives the ground-truth grasp reference point scores and 6D grasp rotations as inputs during training. During inference, these inputs are replaced by the corresponding predictions generated by the first module, thereby establishing the transition between the two stages and enabling end-to-end grasp prediction, as illustrated in Fig.~\ref{fig:DGS-Net}.

Specifically, our network employs PTv3 to extract multi-scale global features $f \in \mathbb{R}^{N \times C}$ from the scene point cloud $\mathcal{P}_k \in \mathbb{R}^{N \times 3}$, where $C$ denotes the feature dimension. Sparse convolution layers are then applied to predict the grasp reference point scores $S_g \in \mathbb{R}^{N \times 2}$ and the 6D rotation representation $R \in \mathbb{R}^{N \times 6}$ for each point. Here, grasp reference point prediction is formulated as a point-wise binary classification task, where the two dimensions of $S_g$ correspond to graspable and non-graspable classes, respectively. After applying the softmax operation, the probability of the graspable class is used as the grasp reference point score.

Based on these, the grasp reference point scores $S_g$, grasp poses $R$, and global point cloud features $f$ are fed into a self-attention encoder followed by multiple sparse convolution layers to predict the pre-grasp joint angles $J_{\text{init}} \in \mathbb{R}^{20}$, the final grasp joint angles $J_{\text{end}} \in \mathbb{R}^{20}$, and the grasp offset $\sigma \in \mathbb{R}^{3}$. The final TCP of the dexterous hand is computed as $T = P_{S_g} + \sigma$, where $P_{S_g} \in \mathbb{R}^{3}$ is selected from the input point cloud $\mathcal{P}_k$ by first filtering points with $\operatorname{Softmax}(S_g)[0] > 0.7$ and then applying spatial non-maximum suppression (NMS). The threshold 0.7 is empirically used to filter low-confidence graspable points and retain more reliable grasp reference points during inference.

Directly regressing high-dimensional joint angles from raw point clouds is challenging due to the weak and ambiguous correlation between geometry and dexterous hand configurations. To address this, we decompose grasp prediction into two stages. A global understanding module first reasons over holistic scene context to identify feasible grasp locations and approach orientations, guided by grasp semantic labels that highlight regions with high grasping potential. A grasp refinement module then exploits local geometric features to predict accurate hand TCP and joint configurations for execution.

In addition, the proposed grasp offset represents the discrepancy between the dexterous hand's tool center and the associated grasp reference point in the observed single-view point cloud. Due to the inherent limitation of single-view perception, the observed point cloud is biased toward visible surfaces. Directly using the observed points as the dexterous hand TCP may therefore lead to finger-object misalignment and unstable contacts. However, this offset is difficult to obtain using predefined geometric rules, since it is closely related to object geometry, occlusion patterns, and surrounding scene context. Therefore, we learn the offset through the network to compensate for the uncertainty introduced by single-view perception. Compared with explicit shape reconstruction, the learned offset avoids reconstruction-induced error accumulation while reducing additional computational overhead.

We also incorporate a pre-grasp action that guides the hand from a collision-safe configuration to the final grasp pose, mitigating grasp failures caused by undesired sequential finger contacts during grasp execution. Together, these designs significantly improve the robustness of dexterous grasping in real-world single-view scenarios.

\textbf{Loss Function:}
For grasp reference point prediction, we introduce grasp point semantic labels, where points in our dataset are annotated as grasp reference points, grasp candidate points, or non-graspable points. Since the number of non-graspable points in the point cloud is overwhelmingly larger than that of grasp reference points, the sample distribution is highly imbalanced. Using a standard cross-entropy loss in this case would easily bias the model toward predicting all points as negative. To address this, we treat grasp candidate points as positive samples and apply weighted cross-entropy loss with category-specific weights (grasp reference points: highest weight $\omega_r = 2$, grasp candidate points: medium weight $\omega_c = 1$, non-graspable points: lowest weight $\omega_n = 0.5$), where grasp candidate points serve as auxiliary positives to balance sample ratios and form spatially coherent ``potentially graspable regions'', thereby enhancing training stability and robustness against sparse annotations. The loss function is defined as follows:
\begin{equation}
\label{equ:1}
\mathcal{L}_{\text{gp}} = -\frac{1}{N} \sum_{i=1}^{N} \omega_i y_i^{\text{gt}} \log \hat{y}_i
\end{equation}
where $N$ denotes the total number of points in the input scene point cloud, $\omega_i$ denotes the category-specific weight for grasp reference points, grasp candidate points, and non-graspable points, $y_i^{\text{gt}}$ is the ground-truth label, and $\hat{y}_i$ is the predicted output of the model.

For grasp rotation regression, inspired by \cite{zhouContinuityRotationRepresentations2019}, we represent 3D rotations using a 6D vector formed by concatenating two orthogonal 3D vectors. Unlike Euler angles or quaternions, this method is continuously differentiable in Euclidean space, avoiding training instability and performance degradation from rotation space topological discontinuities \cite{zhouContinuityRotationRepresentations2019}. In this work, we introduce two loss functions for pose regression: Mean Squared Error (MSE), which measures element-wise errors in the rotation matrix, and Geodesic Rotation Error Loss \cite{mohsenisalehiRealTimeDeepPose2019}, which captures the minimal angular difference between the predicted and ground-truth rotations. These two losses are combined in a weighted manner to balance numerical stability and geometric accuracy.
\begin{equation}
\label{equ:2}
\mathcal{L}_{\text{Mat}} = \frac{1}{n} \sum_{i=1}^{n} \left\| \hat{R}_i - R_i^{\text{gt}} \right\|_F^2
\end{equation}
\begin{equation}
\label{equ:3}
\mathcal{L}_{\text{Geo}} = \frac{1}{n} \sum_{i=1}^{n} \cos^{-1} \left[ 0.5 \left( \operatorname{Tr} \left( \hat{R}_i^\top R_i^{\text{gt}} \right) - 1 \right) \right]
\end{equation}
\begin{equation}
\label{equ:4}
\mathcal{L}_R = (1 - \lambda) \mathcal{L}_{\text{Mat}} + \lambda \mathcal{L}_{\text{Geo}}
\end{equation}
where $n$ denotes the number of grasp reference points, \mbox{$\|\cdot\|_F$} denotes the Frobenius norm, $\hat{R}_i$ represents the predicted rotation matrix, $R_i^{\text{gt}}$ denotes the ground-truth rotation matrix, $\operatorname{Tr}(\cdot)$ is the trace of a matrix, and $\lambda$ is the loss weight.

Due to point cloud discreteness and sensor noise, the highest-scoring grasp reference points often deviate from the dexterous hand's TCP. To address this, our network includes an offset prediction branch to regress the displacement between grasp reference points and the hand's TCP, combining the top-scoring grasp reference point with the predicted offset to compute the dexterous hand's TCP. We use Relative L1 Loss to maintain sensitivity to small-magnitude offsets.
\begin{equation}
\label{equ:5}
\mathcal{L}_T = \frac{1}{n} \sum_{i=1}^{n} \frac{ \left\| \hat{\sigma}_i - \sigma_i^{\text{gt}} \right\| }{ \left\| \sigma_i^{\text{gt}} \right\| + \varepsilon }
\end{equation}

For the prediction of dexterous hand joint configurations, we directly apply the standard MSE loss 
to both the pre-grasp and final joint angles, denoted as $\mathcal{L}_{J_{\text{init}}}$ and $\mathcal{L}_{J_{\text{end}}}$, respectively.

Finally, the total loss is a linear combination of all loss terms:
\begin{equation}
\label{equ:6}
\mathcal{L}_{\text{total}} = \lambda_{\text{gp}} \mathcal{L}_{\text{gp}} + \lambda_{R} \mathcal{L}_{R} + \lambda_{T} \mathcal{L}_{T} + \lambda_{J_{\text{init}}} \mathcal{L}_{J_{\text{init}}} + \lambda_{J_{\text{end}}} \mathcal{L}_{J_{\text{end}}}
\end{equation}
where $\lambda_*$ denotes the weight assigned to each loss component, which are selected via a small-scale validation study to balance the convergence speed and stability of different objectives.

Note that only points labeled as grasp reference points in the semantic labels are associated with valid grasp annotations, including grasp pose, offset, and joint configurations. Therefore, the rotation loss ($\mathcal{L}_{R}$), the offset loss ($\mathcal{L}_{T}$), and the joint configuration losses ($\mathcal{L}_{J_{\text{init}}}$ and $\mathcal{L}_{J_{\text{end}}}$) are computed exclusively for grasp reference points.

\section{Dataset Generation}
\begin{figure*}[!t]
\centering
\includegraphics[width=6.5in]{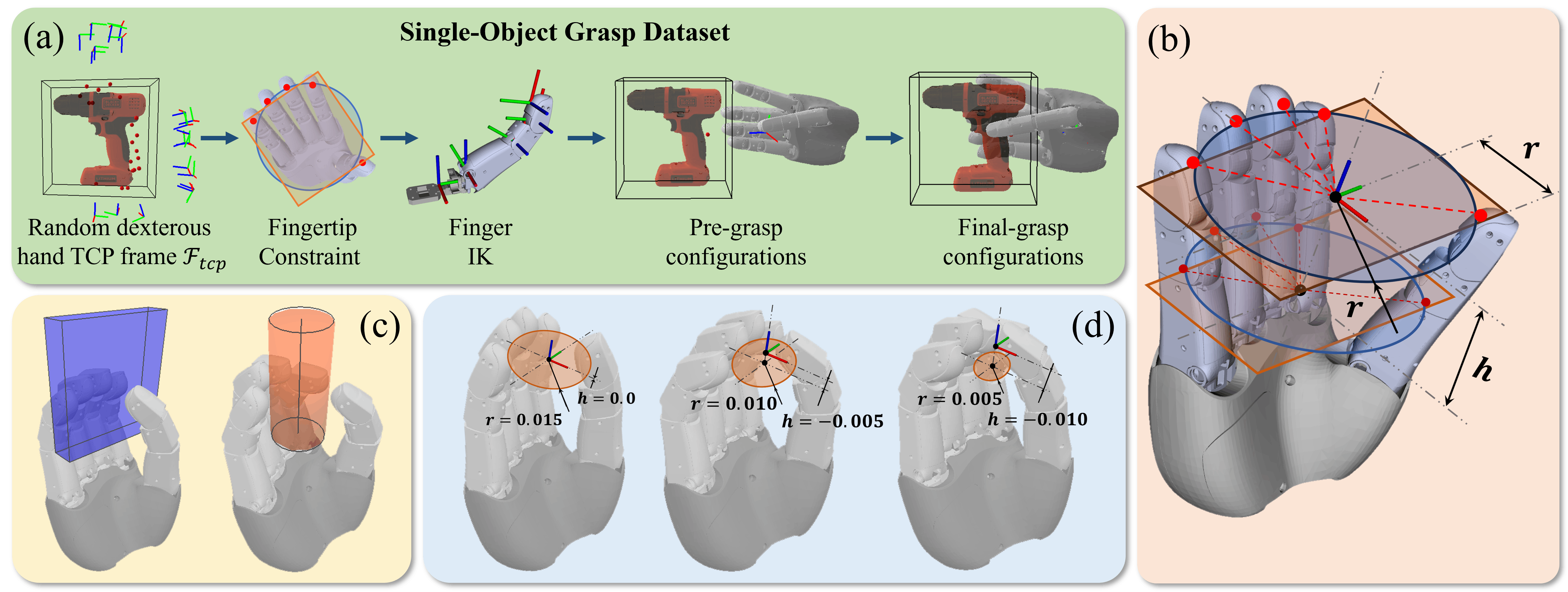}
\caption{(a) Single-object grasp data generation pipeline, generating dense grasps for a single object using fingertip point constraints and admittance control. (b) Circular/rectangular fingertip point constraints. (c) Grasps under different fingertip point constraints. (d) Grasp joint configurations under circular constraints with varying parameters $r$ and $h$.}
\label{fig:object_level}
\end{figure*}

Grasping involves not only hand-object interactions but also environmental and inter-object constraints. Since existing dexterous grasp datasets do not provide grasp offset annotations, we propose a two-stage grasp data generation pipeline that produces not only grasp poses and joint configurations but also point-cloud-level grasp semantic labels and grasp offset annotations. We collect 307 objects from existing datasets and online sources, and apply five scaling factors (0.8, 0.9, 1.0, 1.1, 1.2) to each object to perform data augmentation and enrich grasp diversity across different object sizes. All grasp data are generated in simulation using PyBullet \cite{coumansBulletPhysicsSimulation2015}.

\subsection{Single-Object Grasp Dataset Collection}
Single-object grasp data generation aims to construct a dense and diverse set of physically feasible dexterous grasps for each individual object, which serves as the foundation for subsequent scene-level grasp synthesis. Instead of relying on predefined grasp gestures, we systematically explore the grasp space by jointly sampling grasp positions, orientations, and finger participation patterns, enabling broad coverage of potential dexterous grasp configurations while maintaining physical plausibility.

As illustrated in Fig.~\ref{fig:object_level}(a), each object is first placed at a predefined pose. The tool center point frame $\mathcal{F}_{\mathrm{tcp}}$ of the dexterous hand is randomly placed within the object's bounding box, followed by random rotational perturbations within $[-180^\circ, 180^\circ]$ applied along the $X$, $Y$, and $Z$ axes of the TCP frame $\mathcal{F}_{\mathrm{tcp}}$, resulting in diverse grasp poses. To further increase grasp diversity without explicitly predefining hand gestures, we vary the number of participating fingers and consider three configurations: five-finger, two-finger (thumb and index), and three-finger (thumb, index, and middle).

To obtain feasible grasp configurations from the grasp search space, the dexterous hand is first initialized in a fully open configuration. During closure, circular or rectangular constraints are imposed on the fingertip plane (as illustrated in Fig.~\ref{fig:object_level}(b)) to determine the target positions of each fingertip. These constraints are governed by two parameters: $r$, which denotes the radius of the circle or half the width of the rectangle, and $h$, which represents the distance from the expected positions of the fingertips to the fingertip plane.

\textbf{Circular constraint:} 
\begin{subequations}\label{equ:7}
\begin{align}
\theta_0 = 15^\circ, \quad &\theta_i = 120^\circ + \frac{(i - 1)}{4} \times 120^\circ\label{equ:9a} \\
x_i &= r \cos(\theta_i) \label{equ:7b} \\
y_i &= r \sin(\theta_i)  \label{equ:7c} \\
z_i &= h + \delta_i \label{equ:7d} 
\end{align}
\end{subequations}

\textbf{Rectangular constraint:} 
\begin{subequations}\label{equ:8}
\begin{align}
x_0 = 0.8r, \quad& x_i = -r \label{equ:8a} \\
y_0 = 0.005, \quad&  y_i = 0.025(i-2.5) \label{equ:8b} \\
z_i = h& + \delta_i \label{equ:8c} 
\end{align}
\end{subequations}
Here, the fingertip point $P_i^{\text{tip}} = (x_i, y_i, z_i)$ is expressed in the fingertip plane frame $\mathcal{F}_{\text{tip}}$, where $i = 0, 1, 2, 3, 4$ denotes the finger index, and $\delta_i$ is a finger-specific offset determined by the mechanical structure of the dexterous hand.

The rectangular constraint is suitable for grasping planar objects, while the circular constraint is more appropriate for grasping curved objects such as cylinders and spheres, as illustrated in Fig.~\ref{fig:object_level}(c). To generate more diverse grasp samples, we randomly sample the parameters $r$ and $h$, where different combinations correspond to varying degrees of hand closure (see Fig.~\ref{fig:object_level}(d)). This enables the generation of a richer set of grasp closure configurations.
 
By applying the target constraints, the desired fingertip position for each finger $P_i^{\text{tip}}$ is obtained. This point is then transformed into the local coordinate frame of the corresponding finger $\mathcal{F}_{f_i}$ using the transformation matrix $\mathbf{T}_{\text{tip}}^{f_i}$, resulting in the local position: $P_i^{f_i} = \mathbf{T}_{\text{tip}}^{f_i} \, P_i^{\text{tip}}$. The desired joint angles for each finger, denoted as $\boldsymbol{\Theta}_i = (\theta_{i1}, \theta_{i2}, \theta_{i3}, \theta_{i4})$, are computed via inverse kinematics, where $\theta_{i3}$ and $\theta_{i4}$ are kinematically coupled. During grasp execution, the fingers are actuated using admittance control.

During the admittance control process, the finger is modeled as a mass-damper-spring system, and its motion under external forces is simulated through numerical integration. A relatively low stiffness coefficient $K$ is selected to allow the finger to closely conform to the object surface. The joint angles obtained at the end of the simulation are recorded as the dexterous hand's final grasp configuration $J_{\text{end}}$.

The pre-grasp gesture $J_{\text{init}}$ is derived by scaling the final grasp joint angles $J_{\text{end}}$, with the scaling factors determined empirically. Considering the significant influence of the finger base's yaw and pitch joints on the overall finger posture, a larger scaling factor (0.6) is applied to the pitch joints, while the yaw joints are kept identical to the final grasp configuration (i.e., a scaling factor of 1.0) to preserve the primary grasp posture. For the coupled distal joints, a smaller scaling factor (0.2) is adopted to prevent excessive fingertip closure and premature contact with the object during the approach phase. Here, the pitch and yaw joints refer to rotations about the local $x$- and $z$-axes, respectively. 
\begin{equation}
\label{equ:9}
J_{\text{init}} = \left\{ \theta_{i1},\; 0.6\, \theta_{i2},\; 0.2\, \theta_{i3},\; 0.2\, \theta_{i4} \;\middle|\; \theta_{ij} \in J_{\text{end}} \right\}
\end{equation}

For two-finger and three-finger grasps, the fingers not involved in the grasp retain the same joint configuration in the pre-grasp gesture as in the final grasp gesture. During data collection, the dexterous hand approaches the target object from a distance of 0.1\,$m$ using $J_{\text{init}}$. If any collision occurs with the target during this approach, the grasp attempt is considered a failure. Once the grasp pose is reached, the fingers are closed to the final joint configuration $J_{\text{end}}$ under admittance control, with control parameters chosen to resemble those of real-world dexterous hand executions. A grasp is considered successful if the fingers can effectively envelop the object and maintain stable contact after applying gravity forces along all six axes for 100 simulation steps.
\subsection{Scene-Level Grasp Dataset Collection}

The scene-level grasp dataset is constructed to bridge the gap between isolated object grasp synthesis and realistic multi-object environments, enabling the model to learn grasp feasibility under object-object interactions and environmental constraints. To achieve this, we leverage the single-object grasp sets generated in the first stage and transfer them to multi-object scenes, followed by feasibility validation, which ensures both efficiency and physical consistency.

\begin{figure}[t]
\centering
\includegraphics[width=3.45in]{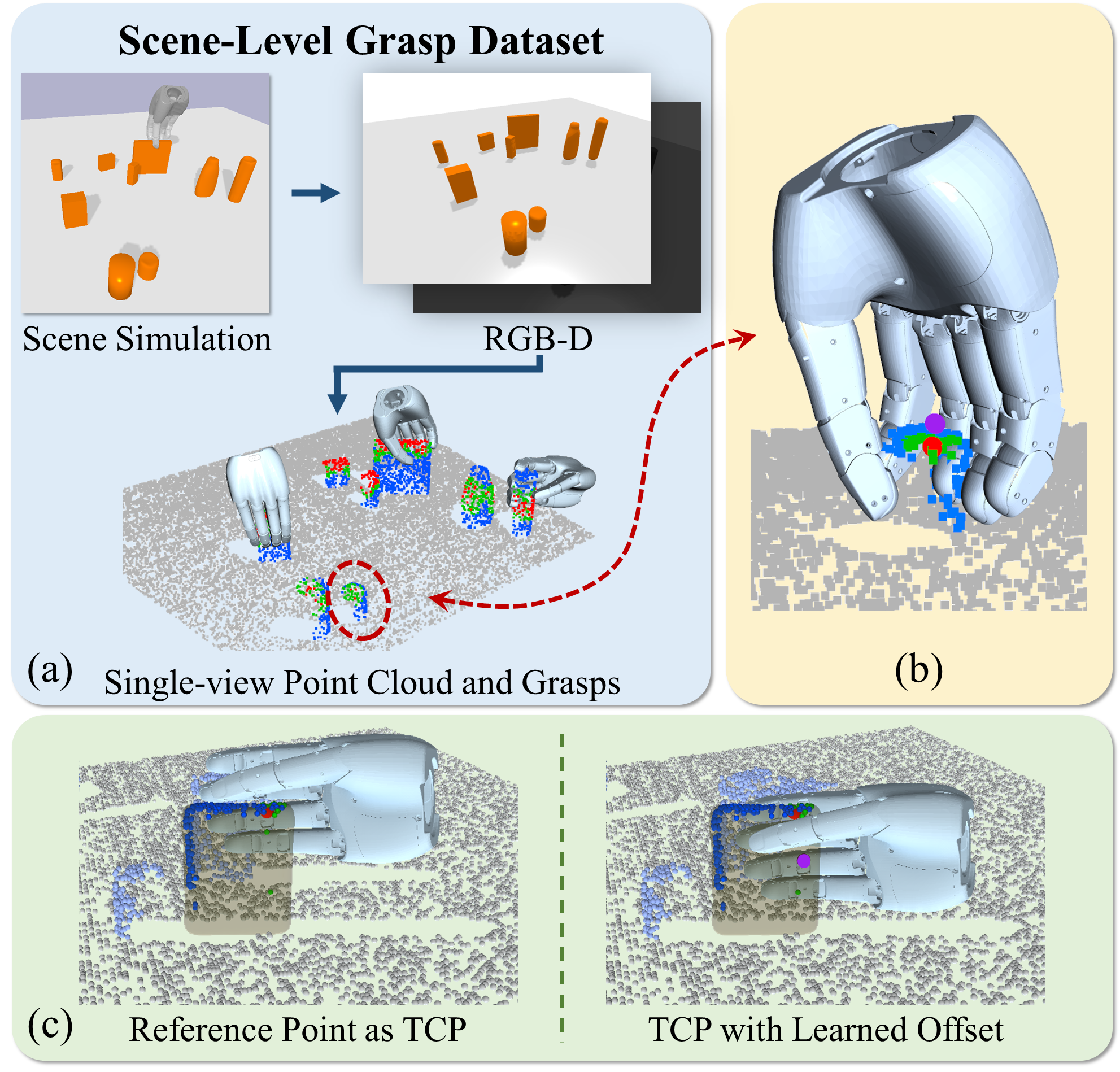}
\caption{(a) Scene-level grasp data generation pipeline. (b) Grasp labels on point cloud: blue represents object point clouds; purple denotes the dexterous hand's TCP $T$; red indicates grasp reference points, which are the points in the object point cloud closest to the hand's TCP; green denotes grasp candidate points in the neighborhood of each grasp reference point; gray point clouds correspond to the tabletop. (c) Illustration of grasp offset compensation.}
\label{fig:scene_level}
\end{figure}

As shown in Fig.~\ref{fig:scene_level}(a), a scene is constructed by randomly sampling 5--15 objects from the first-stage dataset and allowing them to fall under gravity into a stable configuration. The individual-object grasp sets $\mathcal{G} = \{ G_0, G_1, \dots, G_i \}$ are then mapped to the scene, yielding candidate grasps $\tilde{\mathcal{G}} = \{ \mathcal{G}_0, \mathcal{G}_1, \dots, \mathcal{G}_j \}$. Each grasp is validated for collisions, and only those with collision-free approaches are retained as the final scene-level grasp set $\mathcal{G}_{\text{scene}}$. This allows the network to implicitly learn environmental constraints.

Each scene is captured from 256 uniformly sampled viewpoints over a 1-meter-radius hemisphere (elevation: $22.5^\circ$-$67.5^\circ$, azimuth: $0^\circ$-$360^\circ$). For each view, we record the grasp configurations, RGB-D images, segmentation masks, camera extrinsics, and object occlusion relationships.

Notably, the dexterous hand's TCPs generated in Stage~1 are randomly placed within object bounding boxes and do not directly correspond to discrete points in the scene-level point cloud. To generate semantic labels for grasp supervision, we first downsample the scene point cloud and construct a KD-tree for efficient nearest-neighbor search. Each hand's TCP is then associated with its nearest point in the downsampled point cloud, which is designated as the grasp reference point. For each grasp reference point, we further select $0.1N_o$ neighboring points within the corresponding object point cloud as grasp candidate points, where $N_o$ denotes the number of points belonging to that object. This object-adaptive selection strategy adjusts the candidate region according to the visible point-cloud size of each object, providing more consistent annotations under different object scales, sensing conditions, and occlusion levels (see Fig.~\ref{fig:scene_level}(b)). The semantic labels are assigned in a mutually exclusive manner, where grasp reference points take precedence over grasp candidate points in cases of overlap.

Due to single-view point cloud incompleteness, a discrepancy may exist between the dexterous hand's TCP and the corresponding grasp reference point. To address this, we record a grasp-offset target for each instance to supervise the compensation of geometric errors caused by incomplete observations and improve robustness in occluded or unseen regions.

Fig.~\ref{fig:scene_level}(c) illustrates the role of the grasp offset annotation under severe single-view occlusion. If the grasp reference point is directly used as the TCP, the resulting grasp tends to be biased toward the visible side of the point cloud. In this case, the thumb may fail to make effective contact with the object, leading to unstable or failed grasp execution. By introducing the grasp offset, the TCP is shifted from the visible reference point toward a more physically appropriate grasp center, enabling the grasp to account for partially occluded object regions. This annotation improves grasp stability under incomplete observations.

The final training labels consist of:
\begin{itemize}
\item Grasp point semantic labels for point clouds, which could be ``grasp reference point'', ``grasp candidate point'', or ``non-graspable point'';
\item Grasp offset vectors;
\item Dexterous hand grasp poses in the camera frame, along with pre-grasp and final joint configurations.
\end{itemize}

\subsection{Dataset Overview}
Our dataset contains 307 objects spanning five categories (Cuboid, Sphere, Cylinder, Stick, Bowl). The distribution of these object categories, and the grasp distribution of different hand gestures across each category, are presented in Table~\ref{tab:Dataset}. 

\begin{figure}[t]
\centering
\includegraphics[width=3.45in]{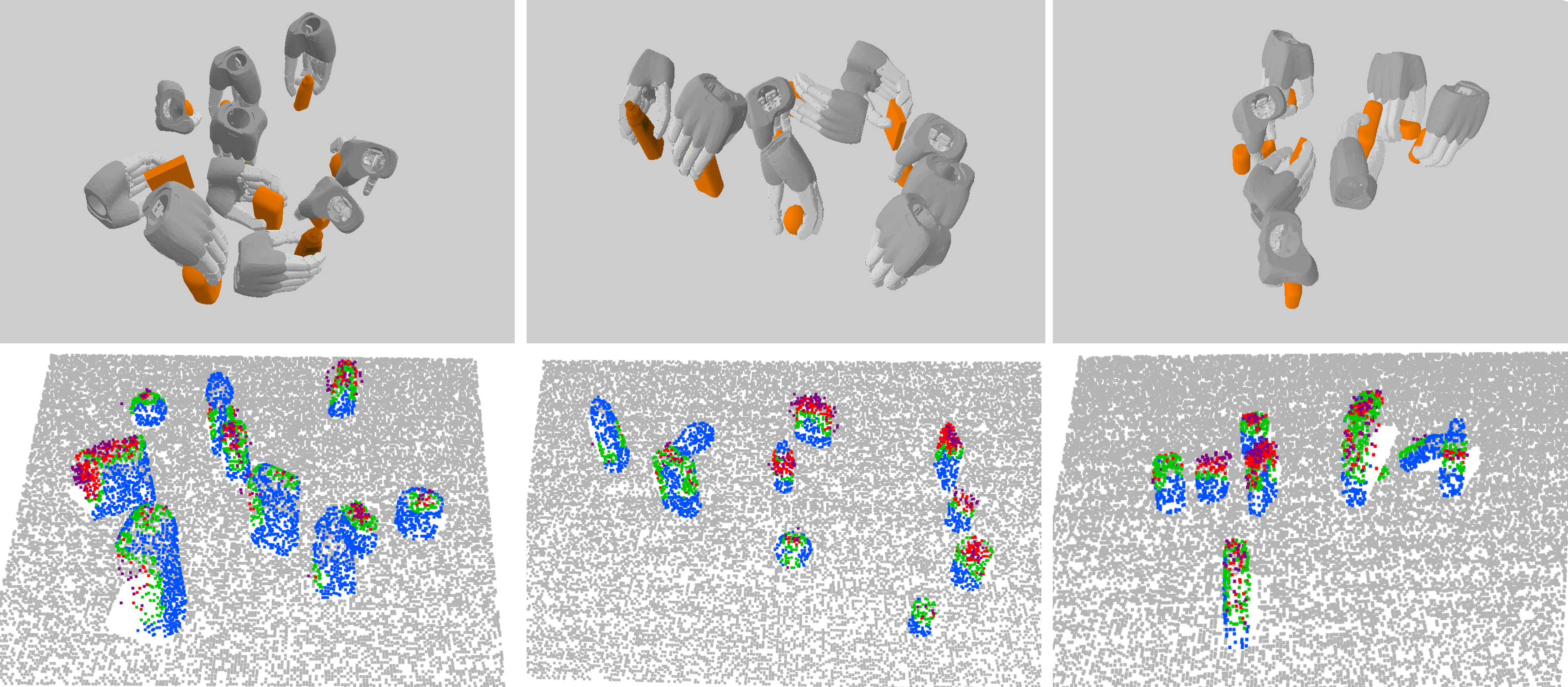}
\caption{Grasp dataset visualization. Top: simulated partial grasps in multi-object scenes. Bottom: single-view point clouds with grasp semantic labels (red: reference points, green: candidate points; blue/gray: non-graspable object/table points) and the dexterous hand TCP (purple) after applying the grasp offset.}
\label{fig:sim_data}
\end{figure}

\begin{table}[t]
\caption{Object and Grasp Data Distribution}
\label{tab:Dataset}
\centering
\begin{tabular}{c|c|ccc}
\Xhline{1pt}
\multirow{2}{*}{ } & \multirow{2}{*}{Object Count} & \multicolumn{3}{c}{Grasps (\%)} \\
\cline{3-5}
                            &                           & 5-finger & 3-finger & 2-finger \\
\hline
Cuboid                      & 116                       & 46.652 & 0     & 0 \\
Cylinder                    & 69                        & 23.147 & 0     & 0 \\
Stick                       & 72                        & 18.442 & 3.366 & 0.231 \\
Sphere                      & 26                        & 4.123  & 2.596 & 0 \\
Bowl                        & 24                        & 1.178  & 0.231 & 0.034 \\
\Xhline{1pt}
\end{tabular}
\end{table}

As illustrated in Fig.~\ref{fig:sim_data}, our dataset contains over 60k RGB-D images from 240 multi-object scenes, each with 5--15 randomly selected objects, captured from 256 viewpoints per scene, yielding more than 350k valid grasp instances. Each sample includes the scene point cloud, grasp point semantic labels, dexterous hand grasp poses, pre-grasp and final joint configurations, occlusion information, and segmentation masks. 
Dataset generation was conducted using PyBullet with multi-process parallel execution on a workstation equipped with an Intel Core i9-14900K CPU. The average generation speeds of Stage~1 and Stage~2 were approximately 3.5 valid object-level grasps/s and 8.0 valid scene-level grasps/s, respectively.

In contrast to datasets that provide only grasp poses and final joint configurations, our dataset additionally includes pre-grasp joint configurations, grasp offsets, and grasp point semantic labels. In our proposed method, grasp candidate points and reference points provided by the semantic labels guide the model to focus on local regions with higher grasp likelihood, while the grasp offsets are used to compensate for geometric and sensing errors introduced by single-view perception, enabling more accurate grasp prediction under occlusion.

\section{Experiments}
In this section, we first present the experimental setup, including evaluation metrics and implementation details. Subsequently, we validate the proposed method in simulation and on a real robotic platform. Experimental results demonstrate that, compared to existing baseline methods, our model can generate more robust grasp candidates, achieving improved grasp success rates and scene clearance rates with lower penetration.
\subsection{Experimental Setup}
\textbf{Evaluation Metrics:} To assess the performance of our proposed method, we adopt several quantitative metrics commonly used in previous works \cite{liHGCNetDeepAnthropomorphic2022,chenDexonomySynthesizingAll2025}:
\begin{itemize}
\item{Grasp Success Rate (SR) (unit: $\%$): The percentage of successful grasps relative to the number of grasp attempts.}
\item{Scene Clearance Rate (CR) (unit: $\%$): The percentage of objects successfully removed relative to the total number of objects initially present in the scene.}
\item{Penetration Depth (PD) (unit: $mm$): The maximum intersection distance between the hand and object for each grasp.}
\item{Penetration Volume (PV) (unit: $mm^3$): The volume of mutual interpenetration between the hand and object for each grasp.}
\item{Diversity (D) (unit: $\%$): The proportion of total variance explained by the first principal component in principal component analysis (PCA), computed as the ratio of the first eigenvalue to the sum of all eigenvalues. PCA is performed on data points that include the hand's TCP $T$, rotation $R$ (in the axis-angle representation), and joint angles $J_{\text{init}}$, $J_{\text{end}}$.}
\item{Grasp Perturbation (GP) (unit: $mm$ and $^{\circ}$): The change in object pose induced by the grasping execution. It is composed of translation error, defined as the Euclidean distance between the object centers before and after grasping, and rotation error, calculated as the angular deviation between the pre-grasp and post-grasp orientations.}
\item{Time Cost (TC) (unit: $s$): The average inference time required for a single grasp prediction.}
\end{itemize}

\textbf{Implementation Details:}
Our network takes a scene point cloud as input, from which 20,000 points are sampled. The model is trained using the Adam optimizer with a learning rate of 0.001 and a batch size of 10 for a total of 40 epochs. Training was conducted on a single NVIDIA RTX 4090 GPU, with an overall training time of approximately 27 hours.

\subsection{Simulation Experiments}
Experiments were conducted in a simulator following this protocol: 240 novel multi-object test scenes are constructed, each containing 10 randomly arranged objects within a $0.6 \times 0.6$\,m workspace. For each scene, single-view point clouds are sampled from random viewpoints on a hemisphere of 1\,m radius centered on the scene. After each model prediction, the grasp with the highest predicted grasp score is executed. The dexterous hand first approaches the target object in the pre-grasp configuration and then executes the final grasp in the final grasp configuration. A grasp is considered successful if the object is lifted at least 10\,cm. After each grasp attempt, regardless of success or failure, the scene point cloud is re-captured and re-evaluated. Successfully grasped objects are removed from the scene, while objects that fail three grasping attempts are also removed. This iterative process continues until no graspable objects remain in the scene.

\textbf{Ablation Experiments:} 
To comprehensively evaluate the effectiveness of the key components in DGS-Net, we conducted the ablation experiments presented in Table~\ref{tab:Ablation}. The full model adopts PTv3~\cite{wuPointTransformerV32024} as the backbone, uses the predicted pre-grasp configuration and grasp offset, and sets the weight of grasp candidate points in $\mathcal{L}_{gp}$ to $\omega_c=1.0$. In the setting without pre-grasp, the network-predicted pre-grasp configurations are replaced with a fixed open-hand configuration ($J_{\text{init}} = \{ \theta_{ij} = 0 \}$). In the setting without offset, the network-predicted grasp offset vectors are discarded and uniformly set to zero. To evaluate the influence of the feature extraction backbone, we replace PTv3 with PointNet++~\cite{qiPointNetDeepHierarchical2017} while preserving the other modules. In addition, to analyze the contribution of grasp semantic labels, we set $\omega_c=0$, which removes the effect of grasp candidate points during training.

\begin{table}[t]
\caption{Ablation Studies in Simulation}
\label{tab:Ablation}
\centering
\footnotesize
\begin{threeparttable}
\begin{tabular}{c|c|cc|c|c|c}
\Xhline{1pt}
\multirow{2}{*}{Offset} & \multirow{2}{*}{Pre-grasp} & \multicolumn{2}{c|}{Backbone} & \multirow{2}{*}{$\omega_c$}& \multirow{2}{*}{SR(\%)} & \multirow{2}{*}{CR(\%)} \\
\cline{3-4}
& & PTv3 & PN++ & & & \\
\hline
\ding{51} & \ding{55} & \ding{51} & & 1.0 & 69.83 & 80.58 \\
\ding{55} & \ding{51} & \ding{51} & & 1.0 & 71.10 & 82.07 \\
\ding{51} & \ding{51} &  & \ding{51} & 1.0 & 69.54 & 81.87 \\
\ding{51} & \ding{51} & \ding{51} & & 0 & 74.38 & 84.97 \\
\ding{51} & \ding{51} & \ding{51} & & 1.0 & \textbf{79.80} & \textbf{87.24} \\
\Xhline{1pt}
\end{tabular}
\begin{tablenotes}
\scriptsize
\item PTv3: Point Transformer V3~\cite{wuPointTransformerV32024}; PN++: PointNet++~\cite{qiPointNetDeepHierarchical2017}. $\omega_c$ denotes the weight assigned to grasp candidate points.
\end{tablenotes}
\end{threeparttable}
\end{table}

Experimental results show that each proposed design contributes to the final grasping performance. Specifically, as the key component for compensating single-view geometric bias, the proposed grasp offset boosts the success rate by 8.70 percentage points (79.80\% vs. 71.10\%). The network-predicted pre-grasp configuration provides a complementary benefit by improving execution robustness, yielding a 9.97 percentage-point improvement in success rate over fixed configurations (79.80\% vs. 69.83\%) and increasing the clearance rate by 6.66 percentage points (87.24\% vs. 80.58\%).

Fixed pre-grasp configurations often result in increased collisions and unstable contacts in multi-object scenes. In contrast, our network-predicted pre-grasp configurations ensure that the fingers are preshaped to match the object geometry, facilitating a coordinated multi-finger envelopment and effectively mitigating object slippage caused by uncoordinated sequential contacts. Furthermore, the proposed grasp offset mechanism proves critical for single-view perception. By explicitly compensating for the geometric centroid shifts induced by partial occlusion and sensor noise, the learned offsets guide the hand toward the true grasp center, thereby significantly enhancing physical consistency and grasp stability.

The choice of backbone also has a clear impact on performance. Compared with PointNet++, PTv3 improves the success rate by 10.26 percentage points (79.80\% vs. 69.54\%) and the clearance rate by 5.37 percentage points (87.24\% vs. 81.87\%). This indicates that the stronger local-global geometric representation learned by PTv3 benefits dense grasp prediction in cluttered scenes. Moreover, using $\omega_c=1.0$ improves the success rate by 5.42 percentage points compared with $\omega_c=0$ (79.80\% vs. 74.38\%). This confirms that emphasizing grasp candidate points helps the network focus on physically meaningful contact regions.

\begin{table}[t]
\caption{Grasp Efficiency Experiment Results}
\label{tab:sim_GE_result}
\centering
\setlength{\tabcolsep}{2pt}

\begin{threeparttable}
\begin{tabular}{c|c|ccc}
\Xhline{1pt}
  & Objects/Scene & SR(\%) &
   CR(\%) & TC($s$) \\
\hline
HGC-Net \cite{liHGCNetDeepAnthropomorphic2022} & 10 & 71.9 & 78.8 & \textbf{0.25} \\
CMG-Net \cite{weiCMGNetEndtoEndContactbased2023} & 10 & 72.0 & 83.3 & 0.59 \\
DGN 2.0 \cite{zhangDexGraspNet20Learning2024} & 10 & 73.59 & 83.87 & 0.5 \\
AnyDexGrasp \cite{fangAnyDexGraspGeneralDexterous2025} & 10 & 74.31 & 85.22 & 0.74 \\
Ours & 10 & \textbf{79.80} & \textbf{87.24} & 0.45 \\
\hline
Multi-FinGAN \cite{lundellMultiFinGANGenerativeCoarseToFine2021} & 1 & 60.0 & -- & 7.4 \\
DVGG \cite{weiDVGGDeepVariational2022} & 1 & 72.4 & -- & 0.9 \\
UniDexGrasp \cite{xuUniDexGraspUniversalRobotic2023} & 1 & 73.7 & -- & -- \\
UniDexGrasp++ \cite{wanUniDexGraspImprovingDexterous2023} & 1 & 85.4 & -- & 0.49 \\
DSQNet \cite{kimDSQNetDeformableModelBased2023} & 1 & \textbf{93.0} & -- & 1.5 \\
Ours & 1 & 92.1 & -- & \textbf{0.45} \\
\hline
Ours & 4--6 & 89.36 & 94.93 & 0.45 \\
\Xhline{1pt}
\end{tabular}
\begin{tablenotes}
\scriptsize
\item DGN 2.0 denotes DexGraspNet 2.0 \cite{zhangDexGraspNet20Learning2024}.
\end{tablenotes}
\end{threeparttable}
\end{table}

\textbf{Grasp Efficiency:} We conduct multi-object grasping experiments in simulation to evaluate the grasp efficiency of DGS-Net. As shown in Table~\ref{tab:sim_GE_result}, our method achieves higher grasp success rates and scene clearance rates compared with the baseline methods. 
Although the runtime is slightly longer (0.45\,s vs. 0.25\,s for HGC-Net), it remains sufficient for scene-level grasping tasks, where prediction speed is not the primary bottleneck. Compared with the state-of-the-art method DexGraspNet~2.0 \cite{zhangDexGraspNet20Learning2024}, our approach achieves comparable performance using a smaller-scale dataset. This indicates that our network learns grasp-relevant representations more efficiently and makes more effective use of training data. In addition, to ensure comparability with real-world grasping experiments, we include an additional set of simulation experiments in which 4--6 objects are randomly placed within a $0.4 \times 0.8$\,m workspace, consistent with the real-world experimental setup.

We further benchmark our approach against representative pipelines that follow an object-centric paradigm, consisting of object segmentation, shape completion, and single-object grasp prediction, including Multi-FinGAN~\cite{lundellMultiFinGANGenerativeCoarseToFine2021}, DVGG~\cite{weiDVGGDeepVariational2022}, and DSQNet~\cite{kimDSQNetDeformableModelBased2023}. As shown in Table~\ref{tab:sim_GE_result}, our method achieves substantially lower inference time, while object-level pipelines can attain slightly higher grasp success rates when operating on isolated single-object inputs. It is worth emphasizing that our approach is capable of directly handling cluttered multi-object scenes from a single-view point cloud, where occlusion is significantly more severe, a setting that remains challenging for object-centric pipelines.

\textbf{Grasp Quality:} We further evaluate grasp quality in simulation using penetration depth, penetration volume, and grasp diversity, as reported in Table~\ref{tab:sim_GQ_result}. Compared with baseline methods, our approach reduces the mean penetration depth by approximately 36.9\% and the mean penetration volume by approximately 61.5\%, indicating more accurate hand-object alignment and improved physical consistency during grasp execution. In addition, the standard deviations of PD and PV are substantially reduced, suggesting that the predicted grasps are not only less penetrative but also more stable and consistent across different scenes.

In terms of grasp diversity, our method performs slightly worse than DexGraspNet 2.0 \cite{zhangDexGraspNet20Learning2024} and Dexonomy \cite{chenDexonomySynthesizingAll2025}. This is mainly because DGS-Net emphasizes physically feasible and stable grasp generation under incomplete single-view observations. In our data generation process, structured contact constraints and physics-based validation are introduced, which help reduce penetration and improve grasp success, but may also narrow the explored grasp space to some extent. In addition, our dataset is smaller than those used by large-scale optimization-based methods, which can further limit grasp diversity. Nevertheless, DGS-Net maintains competitive diversity while achieving superior penetration-related metrics and higher grasp success rates. These results suggest a reasonable trade-off between grasp diversity and physical reliability, demonstrating that DGS-Net can generate robust and physically plausible dexterous grasps in multi-object scenes.

\begin{table}[t]
\caption{Grasp Quality Experiment Results}
\label{tab:sim_GQ_result}
\centering
\setlength{\tabcolsep}{2.5pt}
\begin{tabular}{c|c|c|c}
\Xhline{1pt}
  & PD($mm$) $\downarrow$ & PV($mm^3$) $\downarrow$ & \multirow{2}{*}{D($\%$) $\downarrow$} \\
  & Mean $\pm$ SD & Mean $\pm$ SD & \\
\hline
HGC-Net \cite{liHGCNetDeepAnthropomorphic2022} & 12.93 $\pm$ 5.28 & 13970 $\pm$ 13206 & 51.23 \\
CMG-Net \cite{weiCMGNetEndtoEndContactbased2023} & 13.24 $\pm$ 10.25 & 16524 $\pm$ 25685 & 52.10 \\
GraspTTA \cite{jiangHandObjectContactConsistency2021} & 8.21 $\pm$ 4.63 & 2647.70 $\pm$ 1744.08 & 48.80 \\
DVGG \cite{weiDVGGDeepVariational2022} & 4.10 $\pm$ - & 3800 $\pm$ - & -- \\
UniDexGrasp \cite{xuUniDexGraspUniversalRobotic2023} & 8.24 $\pm$ 6.07 & 16014.49 $\pm$ 14855.84 & 30.27 \\
UniDexGrasp++ \cite{wanUniDexGraspImprovingDexterous2023} & 2.07 $\pm$ 2.17 & 1346.25 $\pm$ 1299.15 & 35.57 \\
Dexonomy \cite{chenDexonomySynthesizingAll2025} & 8.28 $\pm$ 7.76 & 7314.02 $\pm$ 10163.44 & \textbf{25.70} \\
DGN~2.0 \cite{zhangDexGraspNet20Learning2024} & 0.594 $\pm$ 7.13 & 1452.24 $\pm$ 3056.10 & 27.14 \\
Ours & \textbf{0.375 $\pm$ 4.53} & \textbf{559.45 $\pm$ 1262.10} & 35.18 \\
\Xhline{1pt}
\end{tabular}
\end{table}

\subsection{Robot Experiments}

\begin{figure}[b]
\centering
\includegraphics[width=3.4in]{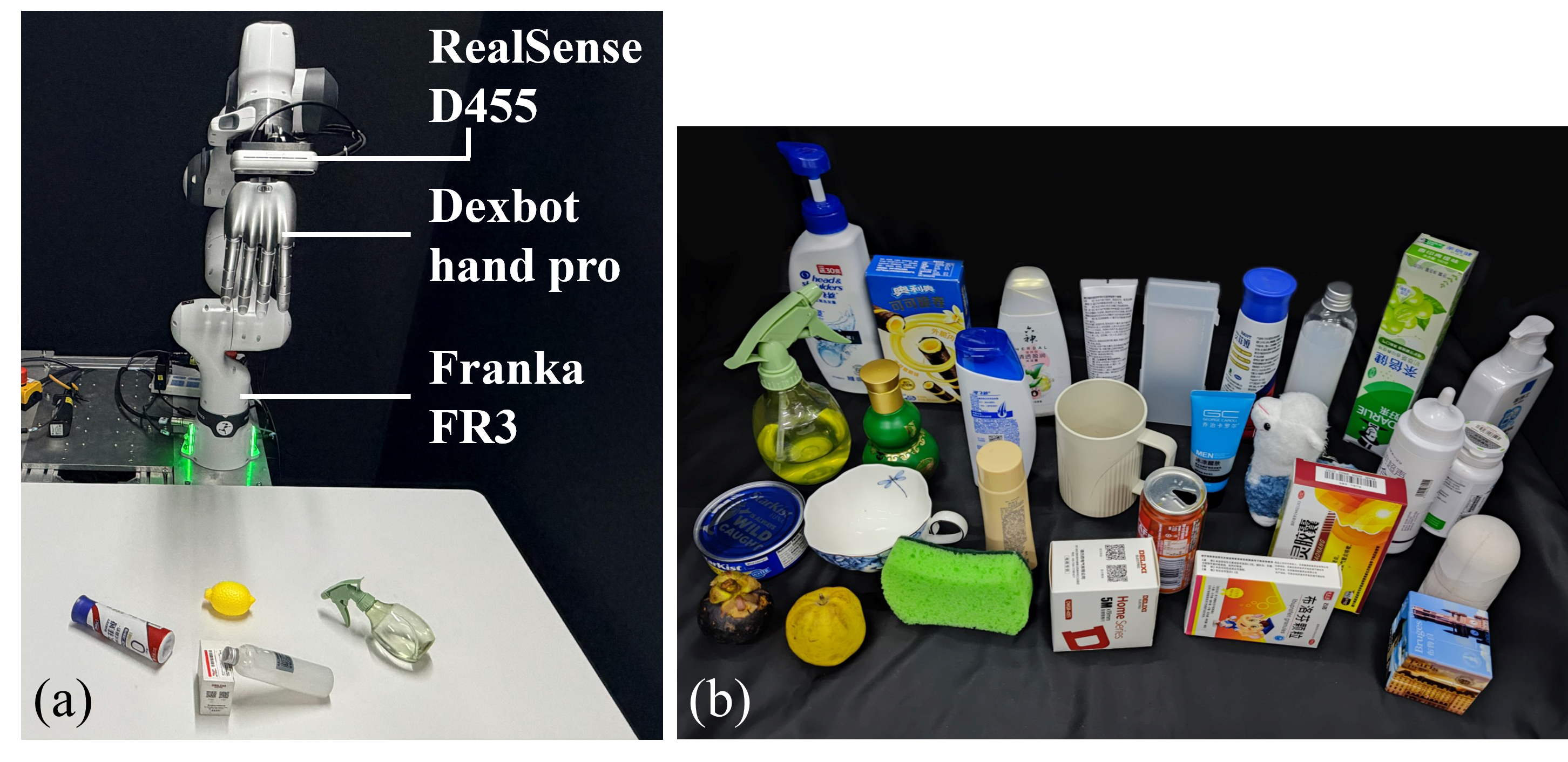}
\caption{Real-world experimental setup: (a) A Franka FR3 equipped with a DexHand Pro and a wrist-mounted depth camera capturing scene point clouds; objects are placed in a $0.4 \times 0.8$\,m workspace. (b) The 29 real-world test objects (unseen during training).}
\label{fig:real_set}
\end{figure}

\begin{figure*}[!t]
\centering
\includegraphics[width=7.14in]{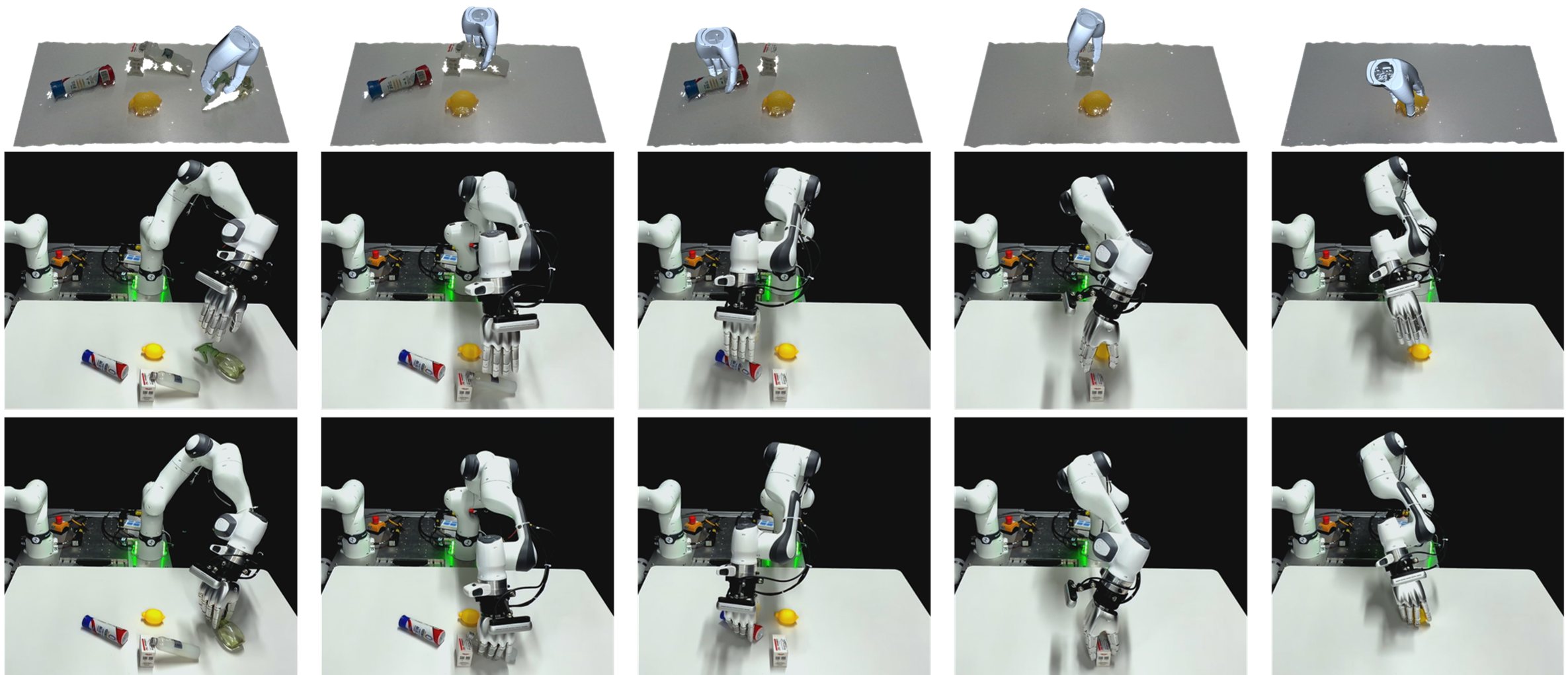}
\caption{Real-world execution pipeline: (top) model prediction, (middle) pre-grasp configuration, and (bottom) final grasp execution.}
\label{fig:real_result}
\end{figure*}

To evaluate the sim-to-real generalization of our method, we deployed a real-world grasping platform consisting of a DexHand Pro mounted on a Franka FR3 robotic arm (Fig.~\ref{fig:real_set}(a)). Scene point clouds were captured by a wrist-mounted RealSense D455 depth camera, and arm motions were planned with MoveIt!. For transparent objects, depth estimation was performed using FoundationStereo~\cite{wenFoundationStereoZeroShotStereo2025}. In all trials, the grasp pose with the highest network-predicted score within the robot workspace was selected, along with its associated offset, pre-grasp configuration, and final grasp configuration. The real-world experiments followed the same protocol as the simulation, with two key differences: (1) each scene contained 4--6 real objects randomly arranged within a $0.4 \times 0.8\,$m workspace, and (2) the object set consisted of 29 novel items (Fig.~\ref{fig:real_set}(b)). Across 30 distinct scenes, we performed a total of 165 grasp attempts, of which 131 grasps were successful, with representative grasping trials shown in Fig.~\ref{fig:real_result}.

As shown in Table~\ref{tab:real_result}, our method achieves a grasp success rate (SR) of 79.39\% and a clearance rate (CR) of 87.94\% in real-world environments, outperforming all baseline methods without any fine-tuning on real-world data.

\begin{figure}[b]
\centering
\includegraphics[width=3.4in]{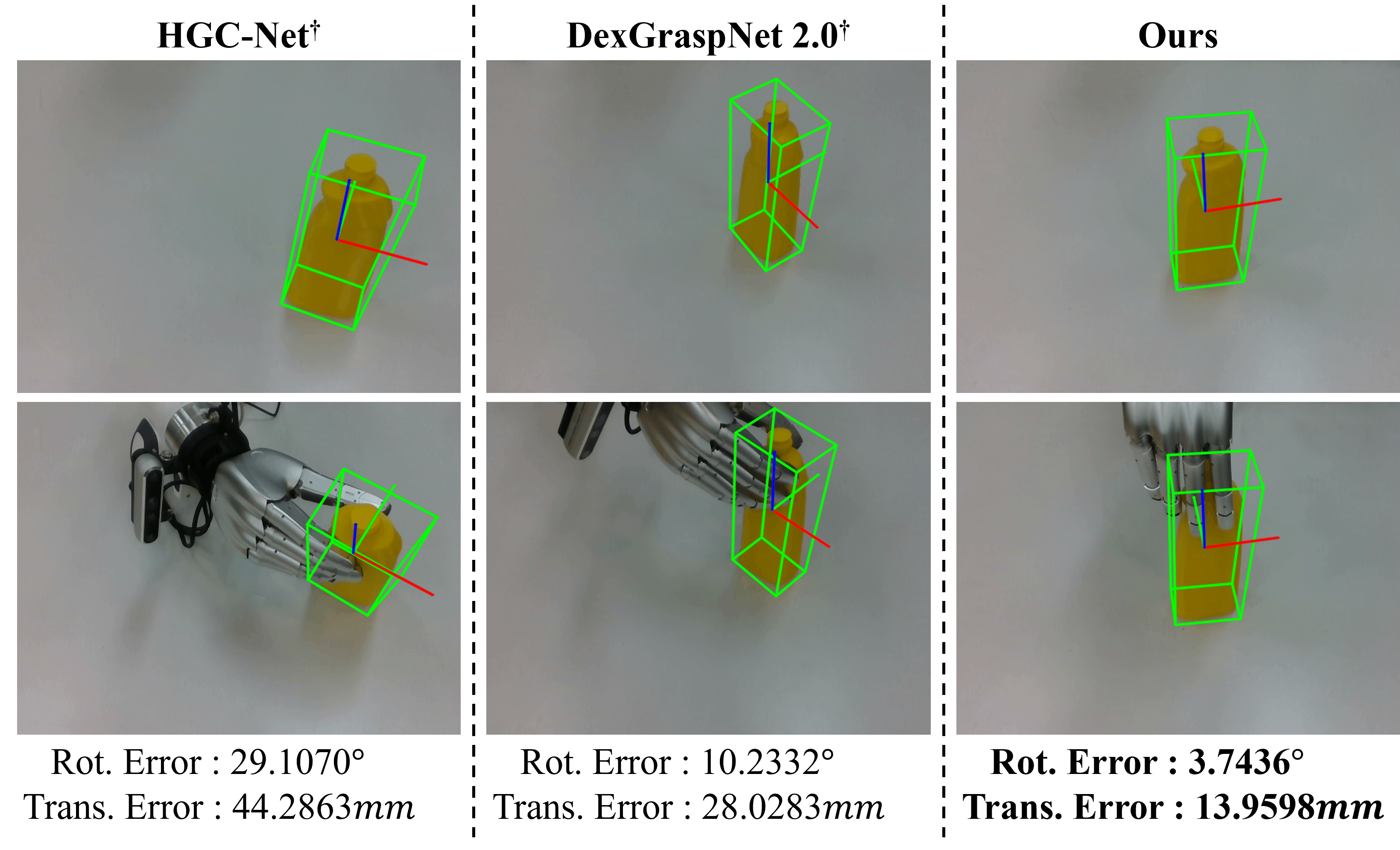}
\caption{Visualization of grasp perturbation.}
\label{fig:real_GP_result}
\end{figure}

To further demonstrate the high quality and precision of the generated grasps, we conducted a grasp perturbation experiment. In this setup, a target object was placed in isolation on the tabletop, and we used FoundationPose~\cite{wenFoundationPoseUnified6D2024} to estimate its 6D pose. We quantified the prediction accuracy by measuring the geometrical deviation of the object induced by the grasping action, specifically the pose variation before and after a successful execution. A minimal pose perturbation indicates a highly accurate grasp, as better alignment minimizes unintended displacement during finger closing. Several examples are illustrated in Fig.~\ref{fig:real_GP_result}.

\begin{table}[t]
\caption{Real-world Experimental Results}
\label{tab:real_result}
\centering
\begin{threeparttable}
\begin{tabular}{c|cccc}
\Xhline{1pt}
    & Ours & DGN~2.0$^\dagger$\cite{zhangDexGraspNet20Learning2024} & HGC-Net$^\dagger$\cite{liHGCNetDeepAnthropomorphic2022} & GraspIt! \\
\hline
SR(\%) & \textbf{79.39} & 74.1 & 66.7 & 58.5 \\
CR(\%) & \textbf{87.94} & 84.5 & 75.0 & 54.3 \\
\Xhline{1pt}
\end{tabular}
\begin{tablenotes}
\scriptsize
\item Modified baseline methods are indicated with $\dagger$.
\end{tablenotes}
\end{threeparttable}
\end{table}

\begin{table}[t]
\caption{Grasp Perturbation Experiment Results}
\label{tab:real_GP_result}
\centering
\begin{tabular}{c|c|c}
\Xhline{1pt}
  & Trans. Error($mm$) $\downarrow$ & Rot. Error($^{\circ}$) $\downarrow$ \\
  & Mean $\pm$ SD & Mean $\pm$ SD \\
\hline
HGC-Net$^\dagger$~\cite{liHGCNetDeepAnthropomorphic2022} & 38.318 $\pm$ 28.337 & 33.979 $\pm$ 25.752 \\
DGN~2.0$^\dagger$~\cite{zhangDexGraspNet20Learning2024} & 17.974 $\pm$ 15.578 & 21.274 $\pm$ 18.045 \\
Ours & \textbf{12.779 $\pm$ 8.317} & \textbf{12.514 $\pm$ 10.106} \\
\Xhline{1pt}
\end{tabular}
\end{table}

We conducted 20 trials for each method, and the results are summarized in Table~\ref{tab:real_GP_result}. Our method substantially outperforms the baselines, achieving the lowest mean errors in both translation (12.779~$mm$) and rotation (12.514$^{\circ}$). Compared to the baseline methods, our method reduces the translation error by approximately 29\% and the rotation error by about 41\%. Furthermore, the substantially lower standard deviations observed in our results indicate that DGS-Net provides more consistent and stable grasping performance.

In summary, extensive experiments in both simulation and the real world validate the effectiveness of our proposed method. In simulation, DGS-Net achieves competitive success rates and clearance rates while substantially reducing penetration depth by 36.9\% and penetration volume by 61.5\% compared to the baselines. These results highlight the efficacy of our proposed two-stage grasp data generation pipeline and network architecture; specifically, the explicit grasp semantic labels provided by our dataset guide the network to focus on grasp-relevant regions, enabling effective learning and robust performance even on a smaller-scale dataset. In real-world robotic experiments, our method maintains robust performance with a 79.39\% success rate and an 87.94\% clearance rate on novel objects. Moreover, the grasp perturbation tests reveal a reduction of approximately 29\% in translation error and 41\% in rotation error, confirming that our explicit modeling of grasp offsets and pre-grasp configurations effectively compensates for single-view perception errors, leading to more accurate and stable dexterous grasping in multi-object scenes.

\section{Conclusion}

This study addresses dexterous grasp prediction in single-view multi-object scenes by proposing an end-to-end grasp prediction network that directly infers grasp positions, grasp poses, and dexterous hand joint configurations from incomplete observations, while also introducing an efficient grasp dataset construction method with point-cloud-level annotations and grasp semantic labels. The proposed grasp offset effectively compensates for the perception deficiencies and geometric uncertainty introduced by single-view observations, leading to more accurate hand-object alignment. In addition, the pre-grasp gesture reduces collisions during grasp execution and improves grasp stability in cluttered multi-object scenes. Extensive simulation and real-world experiments demonstrate that DGS-Net achieves high grasp success and scene clearance rates while consistently producing grasps with lower penetration, validating its robustness and generalization capability under partial observations.

Nevertheless, several limitations remain and will be addressed in future work. First, the current framework mainly focuses on hand-object grasp prediction, while the feasibility of real-world execution is also strongly affected by the robotic arm's kinematic constraints, reachability, and collision-free motion planning. We will therefore couple dexterous grasp prediction with arm trajectory optimization to reduce grasp failures caused by arm motion constraints and improve execution feasibility in cluttered scenes. Second, the current input is limited to visual point clouds and cannot sufficiently capture objects' physical properties such as rigidity, surface roughness, local shape details, density, and mass distribution. Future work will introduce multimodal perception, especially tactile sensing, to estimate these properties and adapt grasp prediction and force control to objects with different materials and physical characteristics. Third, we will further enrich grasp diversity by expanding the dataset and improving the sampling strategy to cover more grasp topologies, particularly two-finger and three-finger grasps, while maintaining physical reliability.

\bibliographystyle{IEEEtran}
\bibliography{refer}

\begin{IEEEbiography}[{\includegraphics[width=1in,height=1.25in,clip,keepaspectratio]{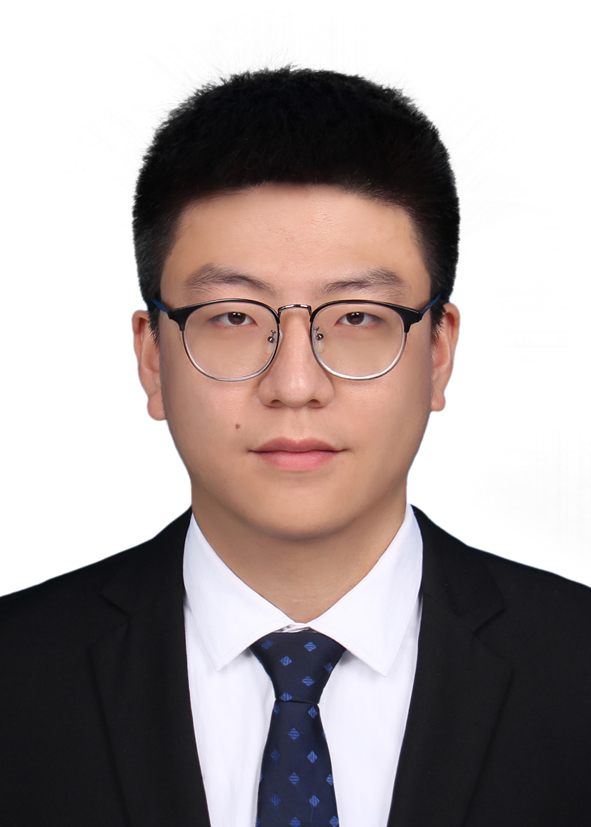}}]{Tao Geng}
received the B.S. degree in mechanical engineering from Harbin Institute of Technology, Harbin, China, in 2023, where he is currently pursuing the Ph.D. degree in mechanical engineering with the State Key Laboratory of Robotics and Systems. \par
His research interests include multifingered dexterous-hand grasping and manipulation, as well as dexterous-hand retargeting.
\end{IEEEbiography}

\begin{IEEEbiography}[{\includegraphics[width=1in,height=1.25in,clip,keepaspectratio]{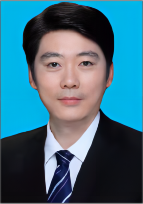}}]{Dapeng Yang}
(Senior Member, IEEE) received the B.S. degree in Mechanical Engineering, and the M.S. and Ph.D. degrees in Mechatronic Engineering from the Harbin Institute of Technology (HIT), in 2004, 2006, and 2011, respectively. \par
He is currently a professor with the School of Mechatronics Engineering, HIT. He is also a member of the State Key Laboratory of Robotics and Systems. His research interests include biomedical signal processing and control, human$\-$machine interaction, and machine learning techniques in robotics.
\end{IEEEbiography}

\begin{IEEEbiography}[{\includegraphics[width=1in,height=1.25in,clip,keepaspectratio]{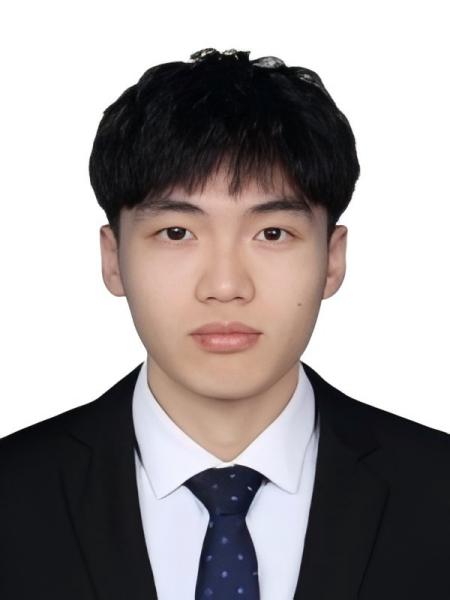}}]{Ziwei Liu}
Ziwei Liu received his B.S. degree in Mechanical Engineering from the School of Mechatronics Engineering, Harbin Institute of Technology, Harbin, China, in 2023, and his M.S. degree in Mechanical Engineering from the State Key Laboratory of Robotics and Systems, Harbin Institute of Technology, Harbin, China, in 2026. \par
His research interests include 6D object pose estimation, 3D object reconstruction, and uncalibrated visual servoing for robotic manipulators.
\end{IEEEbiography}

\begin{IEEEbiography}[{\includegraphics[width=1in,height=1.25in,clip,keepaspectratio]{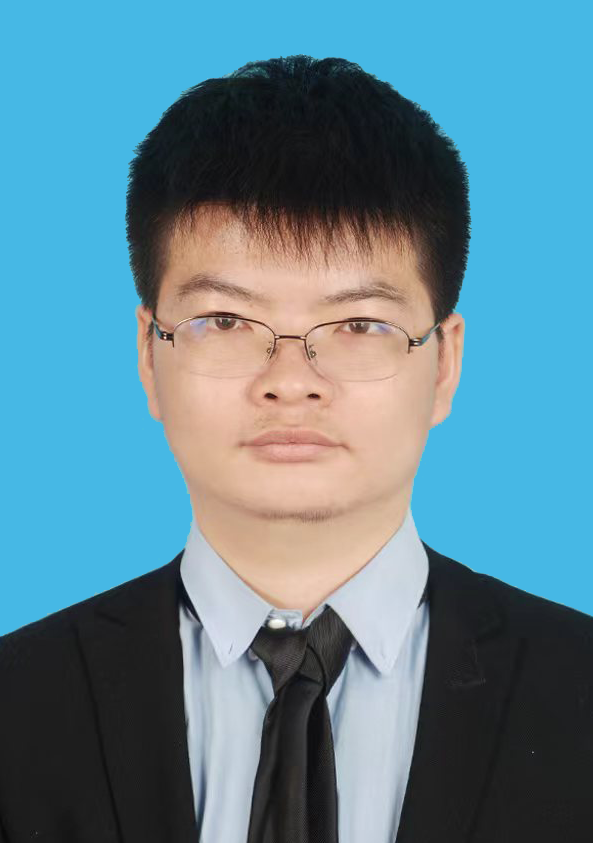}}]{Le Zhang}
received his B.S. degree in Mechanical Engineering from Yanshan University, Qinhuangdao, China, in 2021. \par
He is currently working towards the Ph.D. degree in Mechanical Engineering with the State Key Laboratory of Robotics and Systems, Harbin Institute of Technology, Harbin, China. His research interests include medical robotics, intelligent control technology and soft tissue modeling.
\end{IEEEbiography}

\begin{IEEEbiography}[{\includegraphics[width=1in,height=1.25in,clip,keepaspectratio]{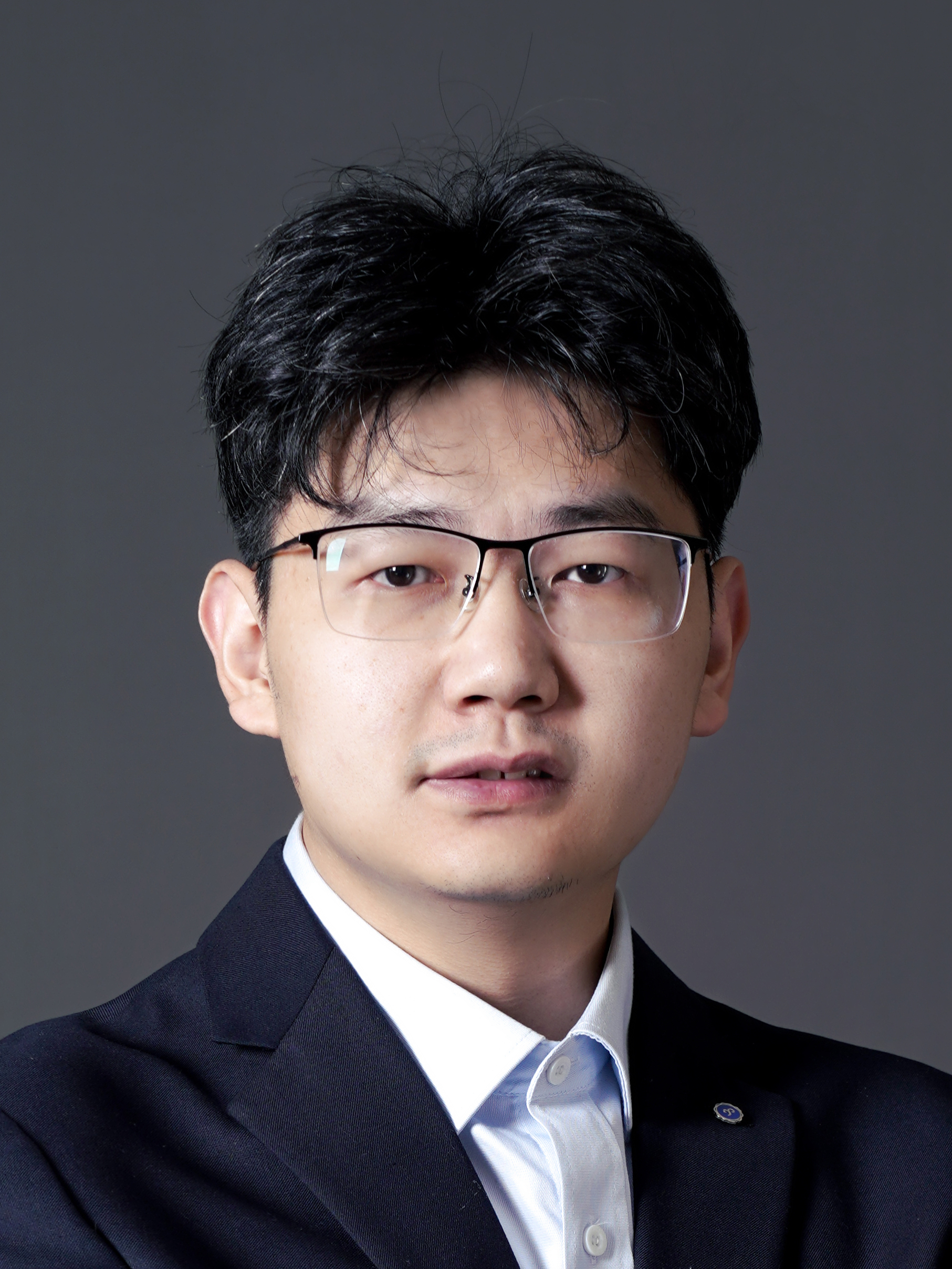}}]{Le Qi}
received the B.S. degree in mechatronic engineering and the Ph.D. degree in mechanical engineering from Harbin Institute of Technology (HIT) in 2019 and 2025, respectively. \par
He is currently an Associate Professor with the School of Mechatronics Engineering, HIT. His research interests include robotic mechanisms and robotic control.
\end{IEEEbiography}

\begin{IEEEbiography}[{\includegraphics[width=1in,height=1.25in,clip,keepaspectratio]{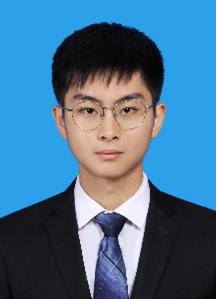}}]{Wangyang Li}
received the B.S. degree in Mechanical Engineering from Harbin Institute of Technology (HIT), Harbin, China, in 2020. He is currently working toward the Ph.D. degree with the State Key Laboratory of Robotics and Systems, HIT. \par
His research interests include the mechatronics design and control algorithm of dexterous robotic hands.
\end{IEEEbiography}

\begin{IEEEbiography}[{\includegraphics[width=1in,height=1.25in,clip,keepaspectratio]{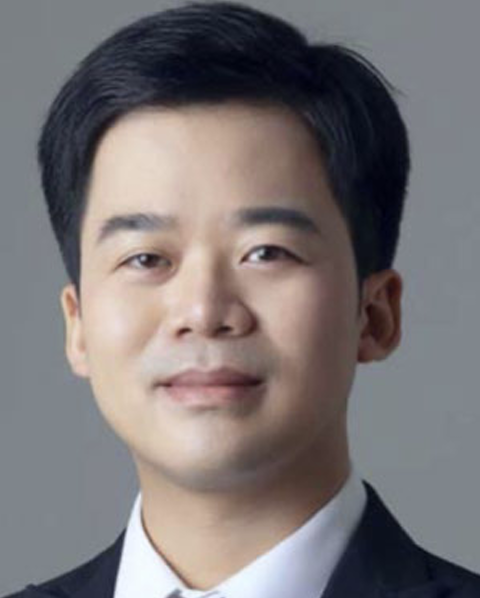}}]{Yi Ren}
(Member, IEEE) received the B.S. degree in thermal energy and power engineering from Harbin Institute of Technology, Harbin, China, in 2011, and the M.Sc. and Ph.D. degrees in mechatronics engineering from the State Key Laboratory of Robotics and Systems, Harbin Institute of Technology, in 2013 and 2017, respectively. \par
From 2018 to 2020, he was a Post-Doctoral Fellow and a Senior Scientist in electrical engineering with the Chair of Information-Oriented Control, Department of Electrical and Computer Engineering, Technical University of Munich, Munich, Germany. From 2020 to 2022, he was a Research Scientist with the Tencent Robotics X Laboratory. In 2023, he joined the Advanced Manufacturing Laboratory, Huawei Technologies, Shenzhen, China, where he is currently a Technical Expert in robotics. His research interests include robotics, nonlinear control, distributed control, perception, and natural language processing with application to multirobot cooperation and manipulation.
\end{IEEEbiography}

\begin{IEEEbiography}[{\includegraphics[width=1in,height=1.25in,clip,keepaspectratio]{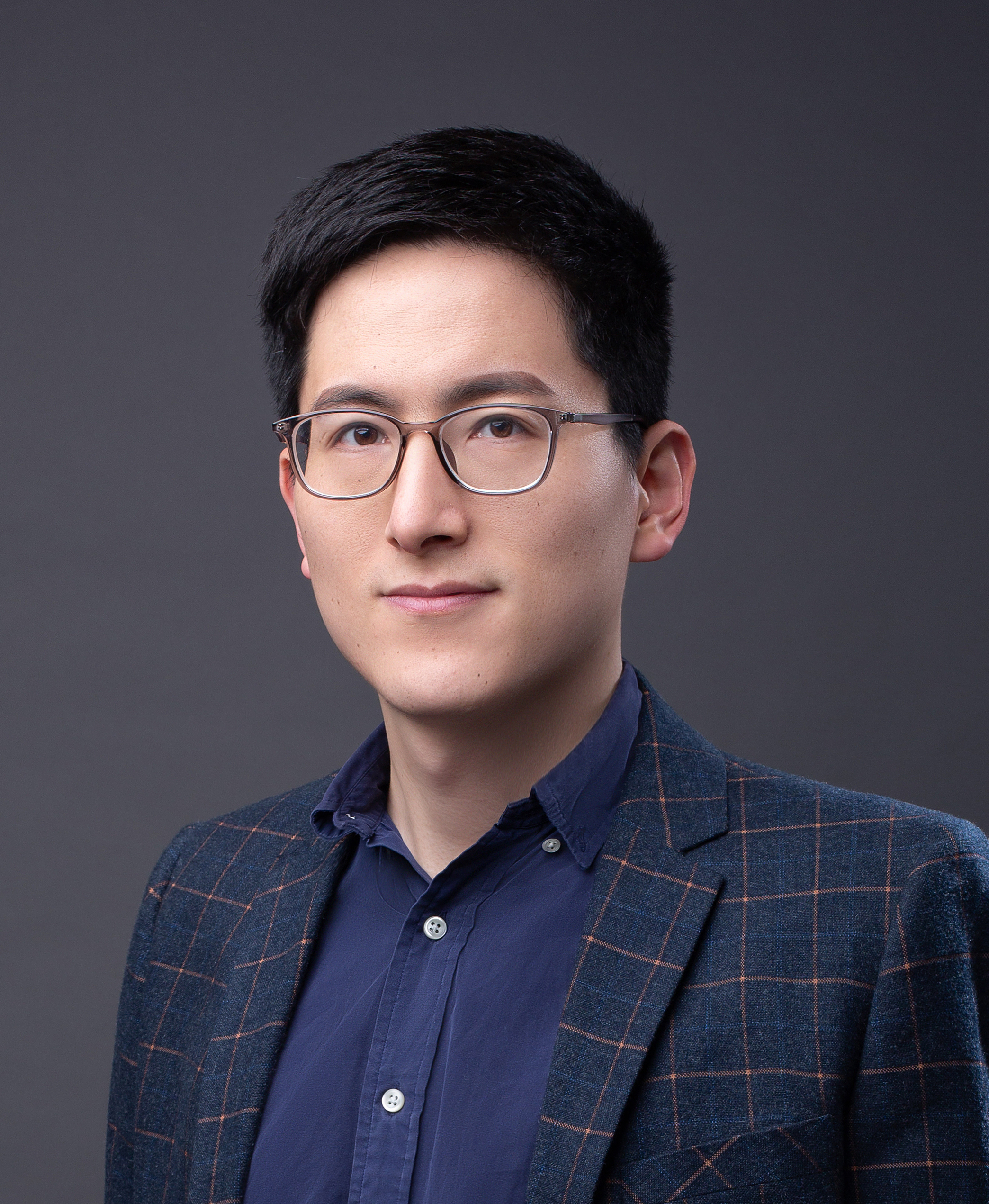}}]{Shan Luo}
(Senior Member, IEEE) received the B.Eng. degree in automatic control from China University of Petroleum, China, in 2012, and the Ph.D. degree in robotics from King's College London, U.K., in 2016. He is currently a Reader (an Associate Professor) in robotics and AI at King's College London. Previously, he was a Lecturer at the University of Liverpool and a Research Fellow at Harvard University and the University of Leeds. \par
His research interests include tactile sensing, robot learning, and visual tactile perception.
\end{IEEEbiography}
\begin{IEEEbiography}[{\includegraphics[width=1in,height=1.25in,clip,keepaspectratio]{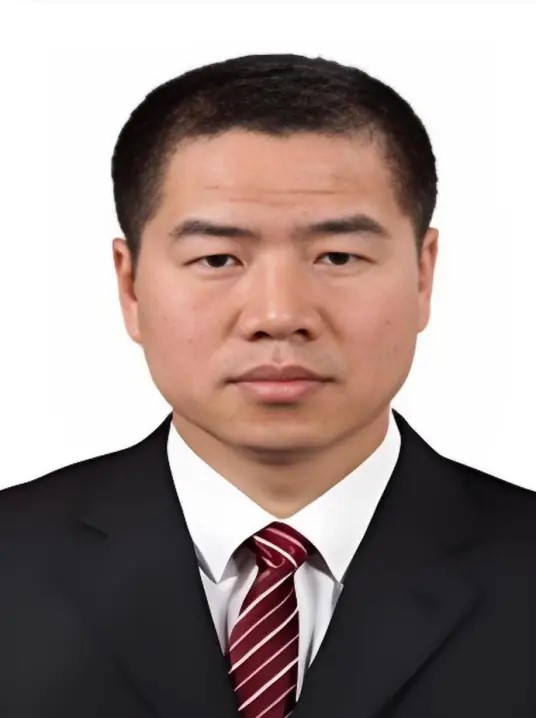}}]{Fenglei Ni}
received the B.S. and M.S. degrees in electrical engineering and the Ph.D. degree in mechanical engineering from Harbin Institute of Technology (HIT), Harbin, China, in 1998, 2002, and 2007, respectively. \par
He is currently a Professor with the State Key Laboratory of Robotics and Systems, HIT. His research interests include AC motor drives, space robotics, and robot control technologies.
\end{IEEEbiography}

\end{document}